%% file: main.tex
\documentclass[runningheads]{llncs}

\usepackage{eccv}

\usepackage{eccvabbrv}
\usepackage{graphicx}
\usepackage{multirow}
\usepackage{floatrow}
\usepackage{amssymb}
\usepackage{pifont}
\usepackage{booktabs}
\usepackage{algorithm}
\usepackage{algpseudocode}
\usepackage{comment}
\usepackage[accsupp]{axessibility}  

\newfloatcommand{capbtabbox}{table}[][\FBwidth]

\usepackage{hyperref}

\usepackage{orcidlink}

\newcommand{\sysname}{{DreamScene}}

\begin{document}
\renewcommand\footnotemark{}
\title{DreamScene: 3D Gaussian-based Text-to-3D Scene Generation via Formation Pattern Sampling}
\titlerunning{\sysname}



\author{Haoran Li\inst{1,2}\orcidlink{0000-0003-2958-2262}\and 
Haolin Shi\inst{1,2}\orcidlink{0009-0003-5587-3706}\and 
Wenli Zhang\inst{1,2} \orcidlink{0009-0002-9564-7917}\and 
Wenjun Wu\inst{1,2} \orcidlink{0009-0009-7590-2571}\and 
Yong Liao\inst{1,2}* \orcidlink{0000-0001-6403-0557}\and  
Lin Wang\inst{3} \orcidlink{0000-0002-7485-4493}\and 
Lik-Hang Lee\inst{4}\orcidlink{0000-0003-1361-1612}\and 
Peng Yuan Zhou\inst{5}*\orcidlink{0000-0002-7909-4059}\thanks{* Corresponding authors} 
}

\authorrunning{H.~Li et al.}



\institute{University of Science and Technology of China \and
CCCD Key Lab of Ministry of Culture and Tourism\\
\email{\{lhr123, mar, zwl384369, wu\_wen\_jun\}@mail.ustc.edu.cn, yliao@ustc.edu.cn} \and 
AI Thrust, HKUST(GZ) and Dept. of Computer Science Eng., HKUST \\ \email{linwang@ust.hk}\and  
The Hong Kong Polytechnic University\\ \email{lik-hang.lee@polyu.edu.hk}\and 
Aarhus University\\ \email{pengyuan.zhou@ece.au.dk} \\
}
\maketitle

\input{sec/0_abstract}
\input{sec/1_intro}
\input{sec/2_related}

\input{sec/6_preliminary}

\input{sec/3_method}

\input{sec/4_experiment}

\input{sec/8_applications}

%
%
\bibliographystyle{splncs04}
\bibliography{egbib}

\newpage
\centerline{\textbf{\Large Supplementary Material}}

\input{sec/9_supplement_arxiv}
\end{document}

%% file: sec/0_abstract.tex

%

\begin{abstract}
Text-to-3D scene generation holds immense potential for the gaming, film, and architecture sectors. 
Despite significant progress, existing methods struggle with maintaining high quality, consistency, and editing flexibility. In this paper, we propose \textbf{\sysname}, a 3D Gaussian-based novel text-to-3D scene generation framework, to tackle the aforementioned three challenges mainly via two strategies. 
First, \sysname\ employs Formation Pattern Sampling (FPS), a multi-timestep sampling strategy guided by the formation patterns of 3D objects, to form fast, semantically rich, and high-quality representations. FPS uses 3D Gaussian filtering for optimization stability, and leverages reconstruction techniques to generate plausible textures. 
Second, \sysname\ employs a progressive three-stage camera sampling strategy, specifically designed for both indoor and outdoor settings, to effectively ensure object and environment integration and scene-wide 3D consistency. 
Last, \sysname\ enhances scene editing flexibility by integrating objects and environments, enabling targeted adjustments. Extensive experiments validate \sysname's superiority over current state-of-the-art techniques, heralding its wide-ranging potential for diverse applications. Code and demos are released
at \url{https://dreamscene-project.github.io}.
 \keywords{Text-to-3D \and Text-to-3D Scene \and 3D Gaussian \and Scene Generation \and Scene Editing}
\end{abstract}

%% file: sec/1_intro.tex
\section{Introduction}
\label{sec:intro}
The advancement of text-based 3D scene generation~\cite{zhang2024text2nerf,hollein2023text2room,ouyang2023text2immersion,cohen2023set,lin2023componerf,hwang2023text2scene,po2023compositional,zhang2023scenewiz3d}  marks a notable evolution in 3D content creation~\cite{poole2022dreamfusion,lin2023magic3d,chen2023fantasia3d,liu2023zero,metzer2023latent,huang2023dreamtime,yu2023text,liang2023luciddreamer,tang2023dreamgaussian,li2023sweetdreamer,nichol2022point,jun2023shap}, broadening its scope from crafting simple objects to constructing detailed, complex scenes directly from text. This leap forward eases 3D modelers' workload and fuels growth in the gaming, film, and architecture sectors.

\begin{figure}[t!]
    \centering
    \includegraphics[width=\textwidth]{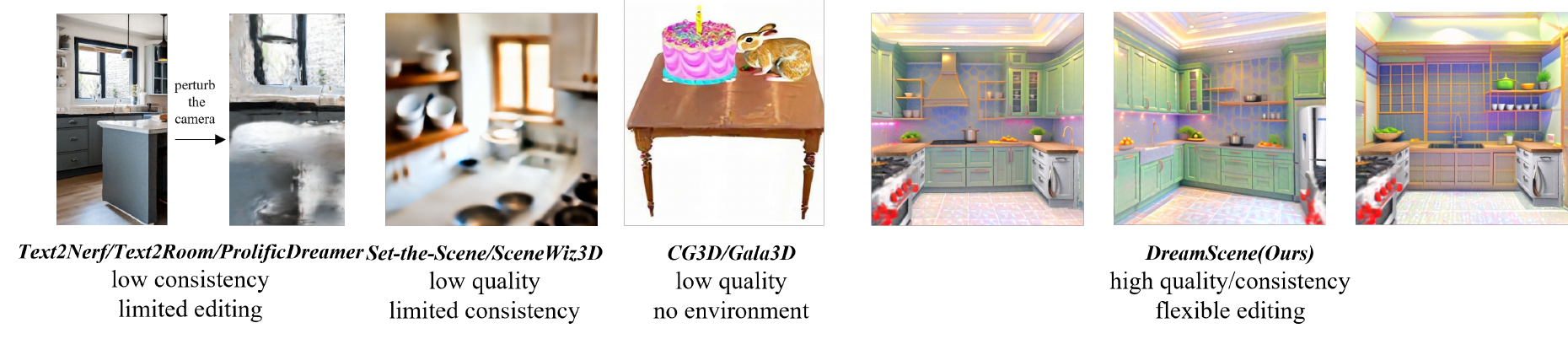}
    \caption{\sysname\ compared with current SOTA text-to-3D scene generation methods. Text2NeRF~\cite{zhang2024text2nerf}, Text2Room~\cite{hollein2023text2room}, and ProlificDreamer~\cite{wang2024prolificdreamer} require $7\sim 12$ hours to generate the scene while \sysname\ only needs \textbf{\textit{1 hour}}. Moreover, \sysname\ is capable of generating scenes that accommodate up to 20 objects as shown in later figures.} 
    \label{fig:compare}
\end{figure}

Text-to-3D methods~\cite{poole2022dreamfusion,lin2023magic3d,chen2023fantasia3d,liu2023zero,metzer2023latent,huang2023dreamtime,yu2023text,liang2023luciddreamer,tang2023dreamgaussian,li2023sweetdreamer,nichol2022point,jun2023shap} typically employ pre-trained 2D text-to-image models~\cite{ramesh2022hierarchical,rombach2022high,saharia2022photorealistic} as prior supervision to generate object-centered 3D differentiable representations~\cite{mildenhall2021nerf,park2019deepsdf,kerbl20233d,muller2022instant,shen2021deep}. While text-to-3D scene generation methods demand rendering from predefined camera positions outward, capturing the scene from these perspectives. These text-to-3D generation methods, as depicted in Fig.~\ref{fig:compare}, encounter several critical challenges, including: \textbf{1)} The inefficient generation process that often leads to low-quality outputs~\cite{zhang2024text2nerf,cohen2023set,lin2023componerf,po2023compositional} and long completion time~\cite{hollein2023text2room,wang2024prolificdreamer}. \textbf{2)} Inconsistent 3D visual cues~\cite{hollein2023text2room,zhang2024text2nerf,cohen2023set,po2023compositional,ouyang2023text2immersion,wang2024prolificdreamer,zhang2023scenewiz3d}, with acceptable results limited to specific camera poses, akin to 360-degree photography. \textbf{3)} Unable to separate objects from the environments, hindering flexible edition on individual elements~\cite{hollein2023text2room,zhang2024text2nerf,ouyang2023text2immersion,wang2024prolificdreamer}.



\par 
In this paper, we introduce \textbf{\textit{\sysname}}, a pioneering 3D Gaussian-based text-to-3D scene generation framework that primarily utilizes the innovative Formation Pattern Sampling (FPS) method. Accompanied by a strategic camera sampling and the seamless object-environment integration, \sysname\ efficiently tackles the challenges above, paving the way for crafting high-quality and consistent scenes. Specifically, based on the observed patterns in 3D representation formation, FPS employs multi-timestep sampling (MTS) to swiftly balance the semantic information and shape consistency in order to generate high-quality and semantically rich 3D representations. FPS ensures stable generation performance by eliminating redundant internal 3D Gaussians during optimization. Finally, employing DDPM~\cite{ho2020denoising} with small timestep sampling and 3D reconstruction techniques~\cite{kerbl20233d}, FPS efficiently generates surfaces with plausible textures from various viewpoints in only \textbf{\textit{tens of seconds}}.

\par 

\par 

Next, we propose an incremental three-stage camera sampling strategy to ensure 3D consistency. First, we generate a coarse environment representation via camera sampling at the scene center. Second, we adapt ground formation to the scene type: a) indoors, by dividing into regions and randomly selecting a camera position for rendering; b) outdoors, by organizing into concentric circles based on the radius, sampling camera poses at varying circles along the same direction. Finally, we consolidate the scene through reconstructive generation in FPS, utilizing all camera poses to further refine the scene.

\par We integrate optimized objects into the scene based on specific layouts to enhance scene generation, thereby preventing the production of duplicate or physically unrealistic artifacts, such as the multi-headed phenomena (generating the same content in all directions) in text-to-3D. Following scene generation, flexible scene editing can be achieved by individually adjusting objects and environments (e.g., modifying positions and changing styles).


Our main contributions can be summarized as follows:
\begin{itemize}
\item We introduce \sysname, a novel framework for text-driven 3D scene generation. Leveraging Formation Pattern Sampling, a strategic camera sampling approach, and seamless object-environment integration, \sysname\ efficiently produces high-quality, scene-wide consistent and editable 3D scenes.
\item Formation Pattern Sampling, central to our approach, harnesses multi-timestep sampling, 3D Gaussian filtering, and reconstructive generation, delivering high-quality, semantically rich 3D representations in 30 minutes.
\item Qualitative and quantitative experiments prove that \sysname\ outperforms existing methods in text-driven 3D object and scene generation, unveiling substantial potential for numerous fields such as gaming and film.
\end{itemize}

%% file: sec/2_related.tex
\section{Related Work}
\subsection{Differentiable 3D Representation}


Differentiable methods such as NeRF~\cite{mildenhall2021nerf,barron2021mip}, SDF~\cite{park2019deepsdf,shen2021deep}, and 3D Gaussian Splatting~\cite{kerbl20233d} enable the representation, manipulation, and rendering of 3D objects and scenes. These representations are compatible with optimization algorithms like gradient descent, facilitating the automatic tuning of 3D representation parameters to reduce loss. Particularly, the recent approach~\cite{kerbl20233d} of modeling 3D scenes with differentiable 3D Gaussians has achieved superior real-time rendering via splatting. Unlike implicit representations~\cite{mildenhall2021nerf,barron2021mip,muller2022instant}, 3D Gaussians provide a more explicit framework, simplifying the integration of multiple scenes. Hence, we adopt 3D Gaussians for their clarity of explicit representation and ease of scene combination.
\subsection{Text-to-3D Generation}

Current text-to-3D tasks mainly generate 3D representation directly~\cite{nichol2022point,jun2023shap,shi2023mvdream} or distilled from large 2D text-to-image models~\cite{poole2022dreamfusion,lin2023magic3d,chen2023fantasia3d}. Direct methods require annotated 3D datasets for rapid generation, often suffering from lower quality and high GPU demands, thus commonly serving as initial steps for distillation techniques~\cite{liang2023luciddreamer,yi2023gaussiandreamer}. For instance, Point-E~\cite{nichol2022point} generates an image via a diffusion model based on text, which is then transformed into the point cloud. Shap-E~\cite{jun2023shap} maps 3D assets to implicit function parameters with an encoder, then trains a conditional diffusion model on these parameters.

Distilling 3D representation from 2D text-to-image models~\cite{poole2022dreamfusion,lin2023magic3d,chen2023fantasia3d,metzer2023latent,huang2023dreamtime,yu2023text} has become the predominant strategy. DreamFusion~\cite{poole2022dreamfusion}, a trailblazer in this domain, introduced Score Distillation Sampling (SDS) to ensure that images rendered from multiple perspectives conform to the distribution of 2D text-to-image models~\cite{ramesh2022hierarchical,rombach2022high,saharia2022photorealistic}. Following its lead, subsequent advancements~\cite{lin2023magic3d,chen2023fantasia3d,liu2023zero,metzer2023latent,huang2023dreamtime,yu2023text,liang2023luciddreamer,tang2023dreamgaussian,li2023sweetdreamer,nichol2022point,jun2023shap} refined 3D generation in terms of quality, speed, and diversity. For example, 
DreamTime~\cite{huang2023dreamtime} accelerates the convergence of generation by monotonically non-increasing sampling of timesteps $t$ in 2D text-to-image model, while LucidDreamer~\cite{liang2023luciddreamer} utilizes DDIM inversion~\cite{mokady2023null,hertz2022prompt}  to ensure 3D consistency of the object generation process. Inspired by these pioneering works, our method introduces a more efficient approach for generating high-quality and semantically rich 3D representation.

\subsection{Text-to-3D Scene Generation Methods}


Current text-to-3D scene generation methods, as illustrated in Fig.~\ref{fig:compare} face significant limitations. \cite{zhang2024text2nerf,hollein2023text2room,ouyang2023text2immersion} relying on inpainted images for scene completion can generate realistic visuals but suffer from limited 3D consistency. Even minor disturbances can lead to the collapse of the scene's visual integrity, indicating a lack of robustness. Moreover, such methods typically do not allow for the editing of objects within the scene and have unrealistic scene compositions, such as a living room cluttered with an excessive number of sofas. 
While some methods~\cite{cohen2023set,zhang2023scenewiz3d}, akin to ours, attempt to merge objects with environments to ensure consistency, they generally yield scenes of lower quality and can accommodate only a limited number of objects. Approaches \cite{vilesov2023cg3d,zhou2024gala3d,lin2023componerf} focusing solely on object integration fall short of generating comprehensive scenes. The constraints on quality, consistency, and editability have thus far hindered the broader application of text-to-3D technologies. Our method represents a leap forward by ensuring high-quality, scene-wide consistency and enabling flexible editing.

%% file: sec/6_preliminary.tex
\section{Preliminary}
\vskip 0.1in \noindent\textbf{Diffusion Models}~\cite{ho2020denoising,song2020denoising} guide data $x$ ($x\sim p(x)$) generation by estimating the gradients of log probability density functions $\nabla_x \log p_{data}(x)$. The training process involves adding noise to the input $x$ over $t$ steps,
\begin{equation}
    x_t = \sqrt{\bar{\alpha_t}}x + \sqrt{1-\bar{\alpha_t}}\epsilon , 
    \label{Eq:ddpm}
\end{equation}
where $\bar{\alpha_t}$ is the predefined coefficient and $ \epsilon \sim \mathcal{N}(0,I)$ is the noise. Then fitting the noise prediction network $\phi$ by minimizing the prediction loss $\mathcal{L}_t$:
\begin{equation}
\mathcal{L}_t = \mathbb{E}_{x, \epsilon \sim \mathcal{N}(0, I)}\left[\left\lVert \epsilon_{\phi}(x_t, t) - \epsilon \right\rVert^2\right] ,
\end{equation}

During the sampling phase, it can deduce $x$ from the noisy input and its noise estimation $\epsilon_{\phi}(z_t, t)$.

\vskip 0.1in \noindent\textbf{Score Distillation Sampling (SDS)} was proposed by DreamFusion~\cite{poole2022dreamfusion} to address the issue of distilling 3D representations from a 2D text-to-image diffusion model. Considering a differentiable 3D representation parameterized by $\theta$ and a rendering function denoted as $g$, the rendered image produced for a given camera pose $c$ can be expressed as $x = g(\theta,c)$. Then SDS distills $\theta$ through a 2D diffusion model $\phi$ with frozen parameters as follows:
\begin{equation}
    \nabla_{\theta}\mathcal{L}_{\text{SDS}}(\theta) = \mathbb{E}_{t,\epsilon,c}\left[w(t)(\epsilon_{\phi}(x_t; y, t) - \epsilon)\frac{\partial g(\theta,c)}{\partial \theta}\right],  
\label{eq:SDS}
\end{equation}
where $w(t)$ represents a weighting function that varies according to the timesteps $t$ and $y$ denotes the text embedding derived from the provided prompt.
\vskip 0.1in \noindent\textbf{Classifier Score Distillation}~\cite{yu2023text} is a variant of SDS inspired by Classifier-Free Guidance (CFG)~\cite{ho2022classifier}, divides the noise difference in SDS into generation prior $\epsilon_{\phi}(x_t; y, t) - \epsilon$ and classifier score $\epsilon_{\phi}(x_t; y, t) - \epsilon_{\phi}(x_t; \emptyset, t)$. It posits that the classifier score is sufficiently effective for text-to-3D, expressed as follows:
\begin{equation}
    \nabla_{\theta}\mathcal{L}_{\text{CSD}}(\theta) = \mathbb{E}_{t,\epsilon,c}\left[w(t)(\epsilon_{\phi}(x_t; y, t) - \epsilon_{\phi}(x_t; \emptyset, t))\frac{\partial g(\theta,c)}{\partial \theta}\right].
\label{eq:CSD}
\end{equation}

\vskip 0.1in \noindent\textbf{3D Gaussian Splatting}~\cite{kerbl20233d, chen2024survey} is a cutting-edge technique in the field of 3D reconstruction. A 3D Gaussian is characterized by a full 3D covariance matrix $\Sigma$ defined in world space centered at point (mean) $\mu$:
\begin{equation}
G(\mathbf{x}) = e^{-\frac{1}{2}\mathbf{x}^T\Sigma^{-1}\mathbf{x}},
\end{equation}
spherical harmonics (SH) coefficients, and opacity $\alpha$. By employing interlaced optimization and density control for these 3D Gaussians, particularly optimizing the anisotropic covariance, one can achieve precise scene representation. The tile-based rendering strategy enables efficient anisotropic splatting, thereby accelerating the training process and achieving real-time rendering capability.

%% file: sec/3_method.tex
\section{Method}
\sysname\ divides the scene into objects and environments. It first generates the objects and then places them within the environments to create the entire scene. This approach is designed to prevent the generation of objects in the environments and to mitigate the multi-head problem in 3D generation (outdoor environments can be generated separately). Specifically, we employ
prompt engineering to develop a large language model (LLM)~\cite{zhao2023survey} agent that transforms scene text into detailed object descriptions $y_i(i=1,..,N)$ and environment description $y_e$, and rapidly create high-quality objects using Formation Pattern Sampling, which consists of multi-timestep sampling, 3D Gaussian filtering and reconstructive generation. Then, we initialize cuboid 3D Gaussians to simulate walls, floors, and ceilings for bounded indoor scenes and hemispherical 3D Gaussians to simulate the ground and faraway surroundings for outdoor unbounded scenes. Through scene layout diagrams, we designate the positions of objects within the scene and use affine transformations (scale $s$, translation $t$, rotation $r$) to place objects in the scene coordinate system. 
\begin{equation}
world(x) = r\cdot s \cdot obj_i(x) + t, i=1,...,N,
\end{equation}
where $x$ denotes the collection of coordinates
for 3D Gaussians. 
Finally, we employ a camera sampling strategy for multi-stage optimization of the environments, achieving scene generation with high 3D consistency. 

\begin{figure}[t]
\centering
\includegraphics[width=\textwidth]{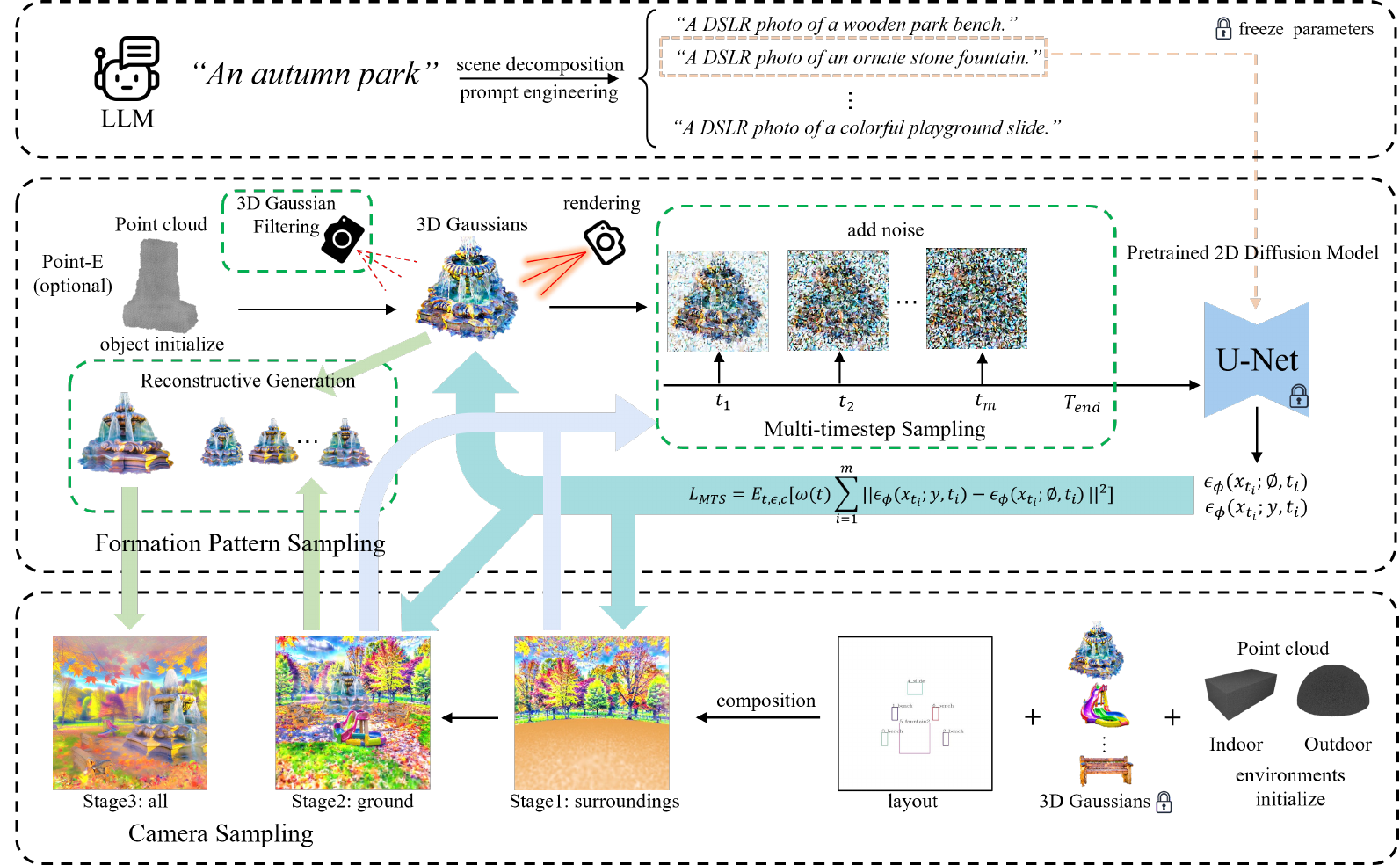}
\caption{ The overview of \sysname. We primarily employ Formation Pattern Sampling, which includes multi-timestep sampling, 3D Gaussian filtering, and reconstructive generation to rapidly produce high-quality and semantically rich 3D representations with plausible textures and low storage demands. Additionally, \sysname\ ensures scene-wide consistency through camera sampling and allows for flexible editing by integrating objects with the environments in the scene.
}
\label{fig:overview}
\end{figure}
\subsection{Formation Pattern Sampling}
\label{sec:Formation Pattern Sampling}

We have refined and broadened the concept of using monotonically non-increasing sampling of timesteps $t$ from DreamTime~\cite{huang2023dreamtime}. We find that the development of high-quality, semantically rich 3D representations benefits from \textbf{\textit{integrating information across sampling multiple timesteps}} of a pre-trained 2D text-to-image diffusion model at each iteration. This finding contrasts with all other methods using SDS~\cite{poole2022dreamfusion}, which typically utilize information from a single timestep at each iteration. Specifically, during the initial and middle phases of optimization, which focus on the primary shape formation, we establish a decremental time window $T_{end}$ that linearly decreases with iterations. $T_{end}$ is segmented into $m$ intervals, within each of which $t$ is randomly sampled and gradients are aggregated. This technique swiftly yields semantically rich 3D representations but may accumulate superfluous 3D Gaussians. To address this, we apply 3D Gaussian filtering to sample only the essential surface Gaussians. In the subsequent stages, focusing on surface texture refinement, we sample $t$ within the range of $0\sim 200$ and employ 3D reconstruction methods~\cite{kerbl20233d} to expedite the generation of plausible textures. In this process, we follow the  pattern of 3D model formation to sample different timesteps 
$t$ at different iterations and 3D Gaussians on the model surface, hence we refer to this process as \textbf{\textit{Formation Pattern Sampling.}} We leverage pseudo-Ground-Truth (pseudo-GT) images, derived from a single denoising step in LucidDreamer~\cite{liang2023luciddreamer}, to encapsulate the diverse information provided by different timesteps $t (0\sim 1000)$ of the 2D text-to-image diffusion model. By adding $t$ timesteps of noise to images $x_0$ to obtain $x_t$, we estimate the pseudo-GT $\hat{x}_{0}^t$ by:
\begin{equation}
\hat{x}_{0}^t = \frac{x_{t} - \sqrt{1 - \bar{\alpha}_t} \epsilon_{\phi}(x_{t}; y, t)}{\sqrt{\bar{\alpha}_t}}.
\label{Eq:pseudoGT}
\end{equation}

\subsubsection{Multi-timestep Sampling.}





\begin{figure}[t]
\centering
\begin{subfigure}{\linewidth}
\centering
\includegraphics[width=1\textwidth]{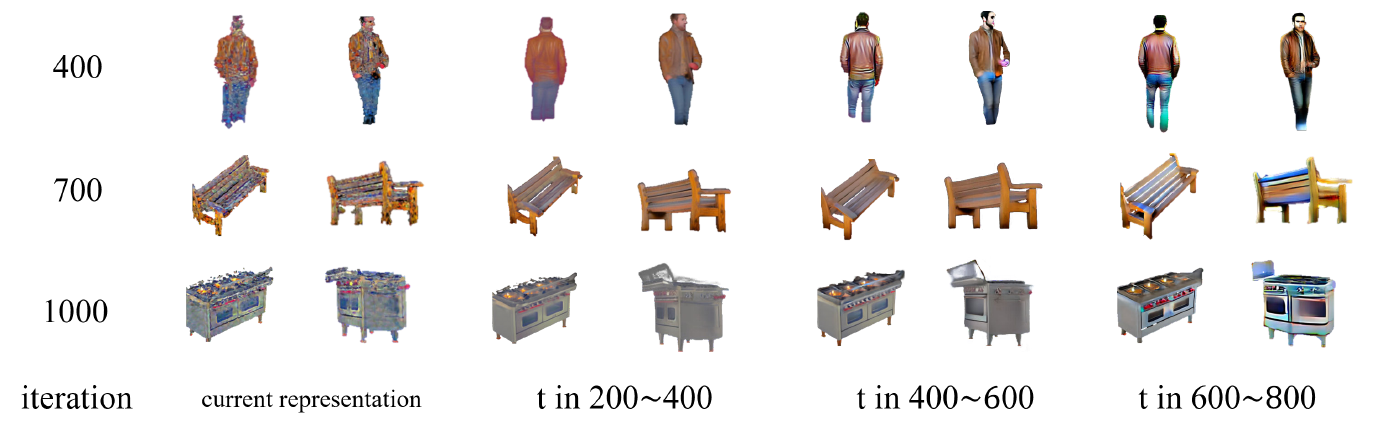}
\caption{ \textbf{Multi-timestep Sampling.} At different timesteps, the 2D text-to-image diffusion model provides different information (represented by the pseudo-GT $\hat{x}_0^t)$ obtained from $x_t$ in a single-step by Eq.~\ref{Eq:pseudoGT} in LucidDreamer~\cite{liang2023luciddreamer}. We demonstrate it by showcasing three objects at various iterations of the optimization process.}
\label{fig:multi-timesteps}
\end{subfigure}
\
\begin{subfigure}{0.48\linewidth}
\centering
\includegraphics[width=\textwidth]{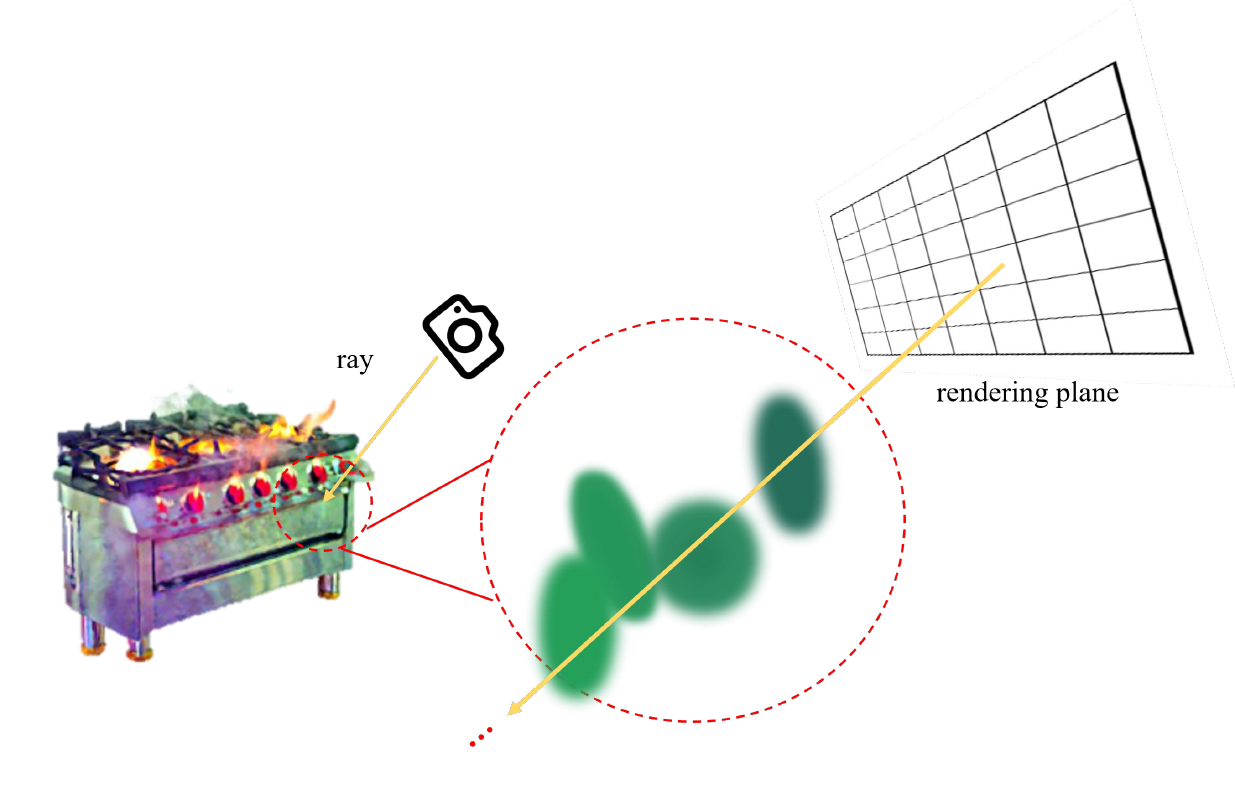}
\caption{ \textbf{3D Gaussian Filtering}. 3D Gaussians that are closer to the rendering plane and have larger volumes contribute more to the rendering.}
\end{subfigure}
\hfill
\begin{subfigure}{0.48\linewidth}
\centering
\includegraphics[width=\textwidth]{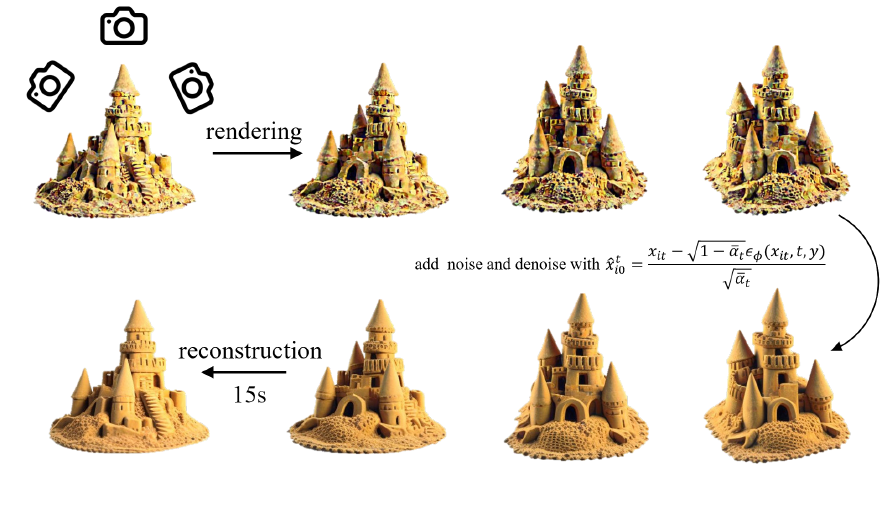}
\caption{ \textbf{Reconstructive Generation.} In the later stages of optimization, generation can be directly achieved through reconstruction based on the denoised images, resulting in 3D representation with fined and plausible textures.}
\label{fig:Gen by Recon}
\end{subfigure}
\caption{Formation Pattern Sampling.}
\label{fig:Formation Pattern Sampling}
\end{figure}

As shown in Fig.~\ref{fig:multi-timesteps}, we discovered when timestep $t$ is small, the 2D diffusion model provides more detailed and realistic surface textures, of which the shape is consistent with the current 3D representation, but lacks in providing rich semantic information of prompt $y$. When $t$ is large, the 2D diffusion model offers abundant semantic information, but its shape may not align with the current 3D representation (e.g., the orientation of the man, the color of the chair, the direction of the cooker between $t$  values of 600 to 800). 
Therefore, we propose to hybridize information from different 2D diffusion timesteps in each iteration, to ensure a certain degree of shape constraint while enriching the semantic information. For instance, in Fig.~\ref{fig:multi-timesteps}, at the $300$-th iteration phase of the man, we need to restrict the shape using the information of timesteps in $200\sim 400$, get richer semantic information
using the information of timesteps in $400\sim 600$ and $600\sim 800$. However, at the $1000$-th iteration of the cooker, we find that the current 3D representation already possesses ample semantic information, and the information provided by larger $t$ might hinder the current optimization direction. Our $i$-th sampled t can be expressed as: 
\begin{equation}
    t_i = T_{end}^{iter}\cdot random(\frac{i-1}{m},\frac{i}{m}),i=1,...,m,
\end{equation}
where  $T_{end}$ denotes a linear decreasingly time window similar to DreamTime, $iter$ represents the current iteration, and $m$ denotes the number of intervals. We adopt DDIM Inversion to sample from $t_1$ to $t_m$ to ensure content consistency~\cite{liang2023luciddreamer}.
\par Therefore, the multi-timestep sampling (MTS) combined with the CSD (Classifier Score Distillation)~\cite{yu2023text} method can be expressed as:
\begin{equation}
     \nabla_{\theta}\mathcal{L}_{\text{MTS}}(\theta) = \mathbb{E}_{t,\epsilon,c}\left[\sum\limits_{i=1}^{m} w(t_i)(\epsilon_{\phi}(x_{t_i}; y, t_i) - \epsilon_{\phi}(x_{t_i}; \emptyset, t_i))\frac{\partial g(\theta,c)}{\partial \theta}\right].
\label{equ:MTS}
\end{equation}

\subsubsection{3D Gaussian Filtering.}
Excessive Gaussians can hinder the optimization process. Unlike current methods  \cite{fan2023lightgaussian,lee2023compact} that filter reconstructed 3D Gaussians using ground truth images, our approach necessitates filtering during optimization. In terms of rendering, the 3D Gaussians closer to the rendering plane inherently have a more significant impact, for which we employ a score function tailored to quantify such contributions. For 3D Gaussians along the rendering ray $r_j$, their contributions are calculated using the inverse square of their distance to the rendering plane, adjusted by the 3D Gaussians volume. This method prioritizes 3D Gaussians that are both closer to the rendering plane and larger in volume as shown in Fig.~\ref{fig:faux_renderding}. By ranking various viewpoints according to scores, we can efficiently eliminate 3D Gaussians falling below a designated threshold. 
\begin{equation}
Score(i) = \sum_{j=1}^{H\times W \times M} \frac{V(i)}{D(r_j,i)^2\times maxV(r_j)},
\end{equation}
where $H$ and $W$ denote the height and width of the rendered image, $M$ indicates the number of rendered images, $V(i)$ represents the volume of the $i$-th 3D Gaussian (calculated using the covariance matrix), $maxV(r_j)$ represents the maximum volume of the 3D Gaussians on $r_j$, and $D(r_j,i)$ is the distance of the $i$-th 3D Gaussian from the rendering plane along the $r_j$. Note that this procedure mirrors the rendering process instead of performing the actual rendering.

\subsubsection{Reconstructive Generation.}
Next, we can expedite the generation of plausible surface textures using 3D reconstruction methods~\cite{kerbl20233d}. During the previous process, we find that when sampling very small timesteps $t$($0\sim 200$), the image $\hat{x}_{0}^t$ predicted by Eq.~\ref{Eq:pseudoGT} has no difference in 3D shape compared to the image $x_0$ before noise addition, but offers more detailed and plausible textures. Therefore, under the premise of shape consistency, we can directly \textbf{\textit{generate}} a new 3D representation through \textbf{\textit{3D reconstruction}}~\cite{kerbl20233d}. As shown in Fig.~\ref{fig:Gen by Recon}, after obtaining a coarse texture and highly consistent 3D representation, we render $K$ images $x_i,i=1,...,K$ from various camera poses $c_i$ around the 3D representation. By adding $t$ timesteps of noise to the images to get $x_{it}$ by Eq.~\ref{Eq:ddpm}, we estimate the images $\hat{x}_{i0}^t$ with  plausible textures using Eq.~\ref{Eq:pseudoGT}, and then reconstruct them on the coarse representation through \cite{kerbl20233d} by minimizing the following reconstruction loss:
\begin{equation}
L_{rec} = \sum_i ||g(\theta , c_i) - \hat{x}_{i0}^t||_2 .
\end{equation}

This process generates a representation with detailed and plausible textures within \textbf{\textit{15 seconds}}.

\subsection{Camera Sampling}
To ensure generation quality, current methods~\cite{hwang2023text2scene,zhang2024text2nerf,ouyang2023text2immersion,cohen2023set,wang2024prolificdreamer} often conduct camera sampling within a limited range, failing to guarantee scene-wide observations. Straightforward random camera sampling within the scene can cause scene generation to collapse during optimization. Thus, we propose an incremental three-stage camera sampling strategy:

\par In the first stage, we generate a coarse representation of the surroundings (indoor walls and outdoor faraway surroundings). We freeze the 3D Gaussians parameters of the ground and objects, sampling camera coordinates within a certain range of the center to optimize the generation of the surroundings.


\par The second stage focuses on generating the coarse ground. We freeze the 3D Gaussians parameters of the environments and objects. The indoor scene is divided into regions based on the placement of objects, we sample camera poses pointing at these focal regions containing objects and the ground of the room in each iteration. The outdoor scene is divided into concentric circles based on radius, and in each iteration, a same direction is chosen to sample camera poses on different circles to optimize ground generation. 
Our strategy is able to cover the entire ground as uniformly as possible and focuses on optimizing the parts where the ground is in contact with objects and the surroundings.





\par In the third stage, we use all the camera poses above to ensure a scene-wide view and refine all environmental elements. This includes optimizing parameters for both the ground and the surrounding environments. Based on the 3D consistency of the previous phases, we then proceed to the reconstructive generation method to obtain more detailed and plausible textures. 


\par Generated camera positions may be obstructed by placed objects, necessitating collision detection between the camera and objects. If a collision occurs, the generated camera should be discarded. Due to space constraints, we have placed the details on the camera sampling strategy in the Supplementary Materials.

%% file: sec/4_experiment.tex
\section{Experiment}
\begin{figure}
\centering
\begin{subfigure}{\linewidth}
\centering
\includegraphics[width=1\textwidth]{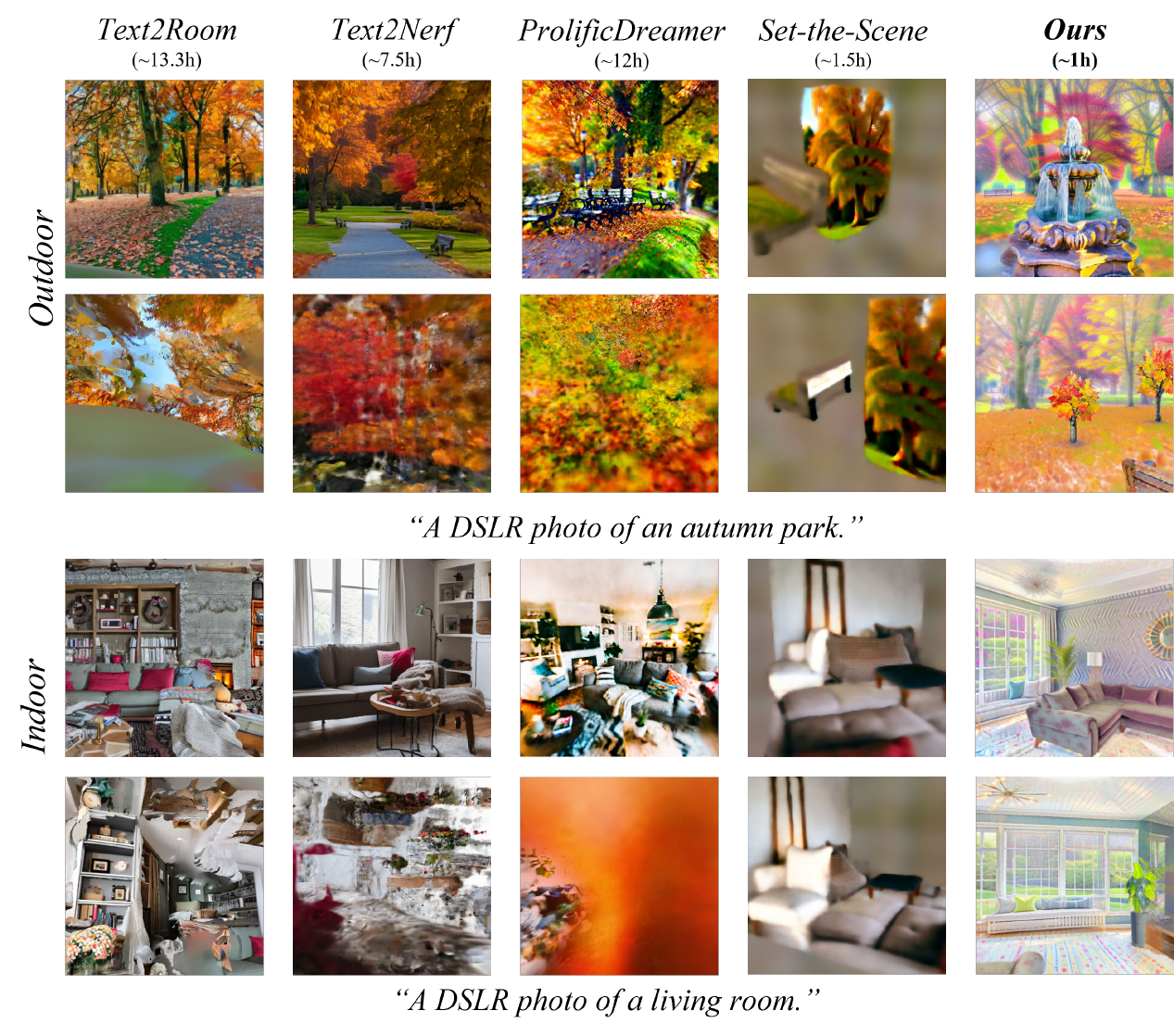}
\caption{Scene Results. The upper part shows images rendered using camera poses employed during the generation process, whereas the lower part shows the images rendered with the randomly selected camera poses within the scene.
}
\label{fig:scene_result}
\end{subfigure}

\begin{subfigure}{\linewidth}
\centering
\includegraphics[width=1\textwidth]{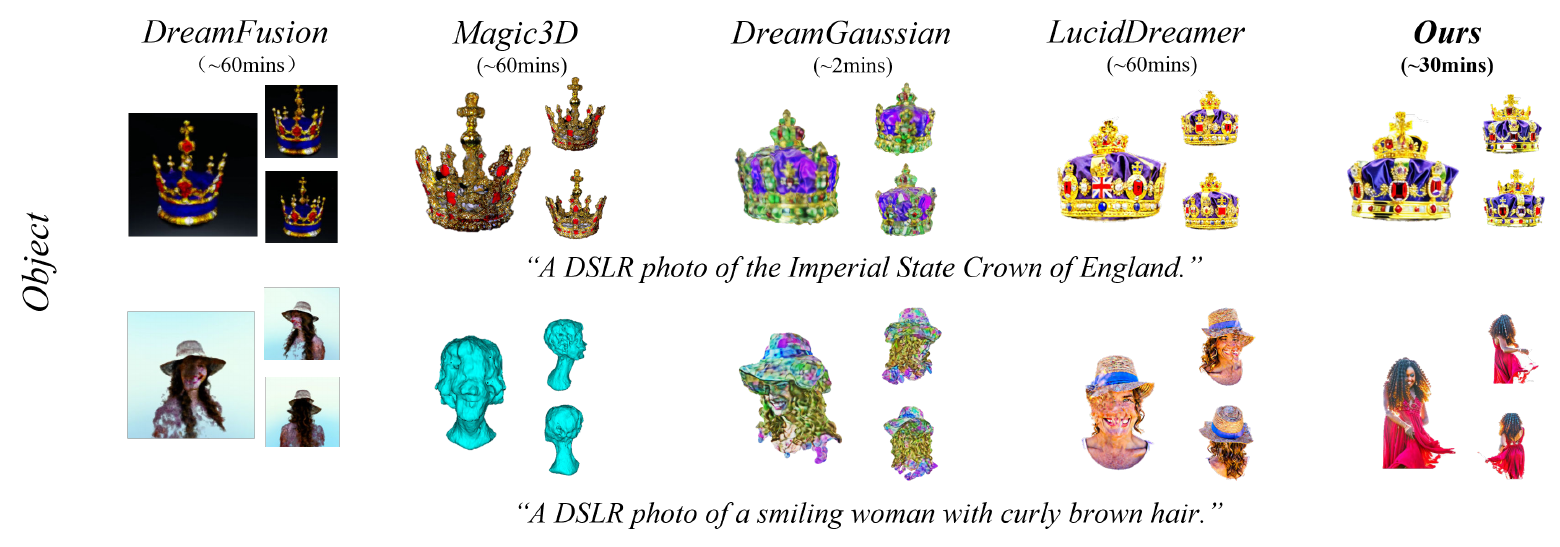}
\caption{ Object Results.}
\label{fig:object_result}
\end{subfigure}
\caption{Comparison with baselines in text-to-3D generation.}
\label{fig:main_result}
\end{figure}

\begin{figure}
\centering
\includegraphics[width=1\textwidth]{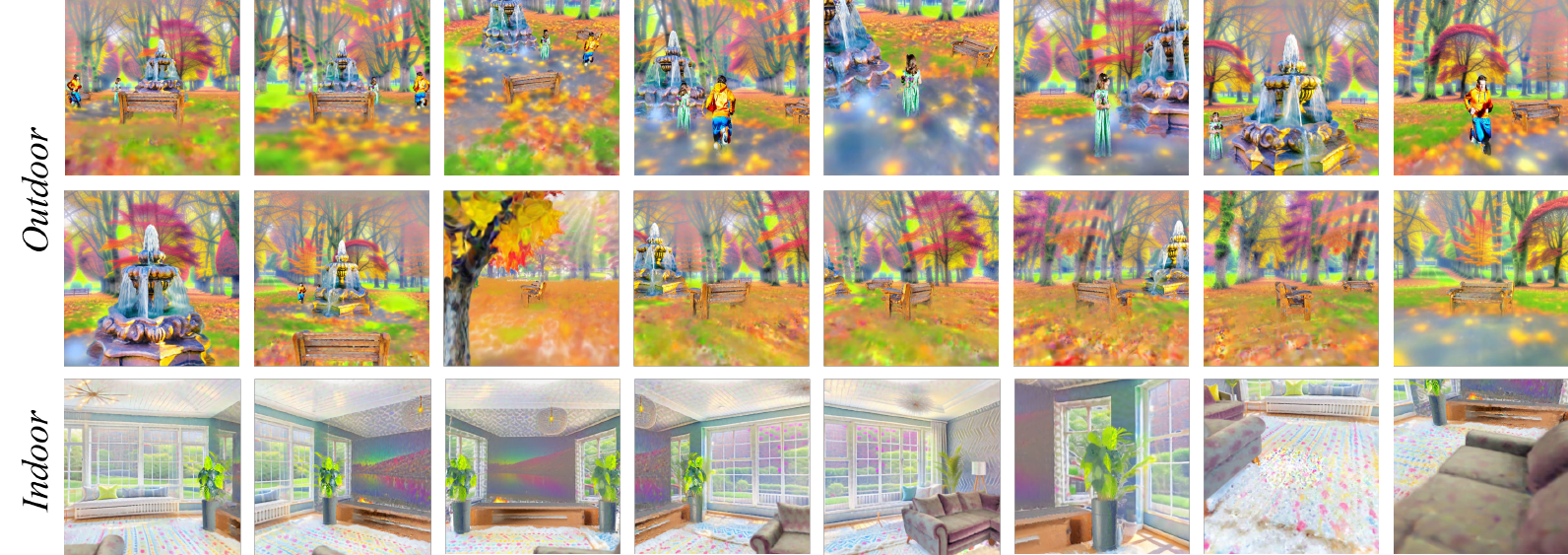}
\caption{Consistency results under multiple scene-wide camera poses.}
\label{fig:consistency}
\end{figure}

\begin{figure}
\centering
\includegraphics[width=1\textwidth]{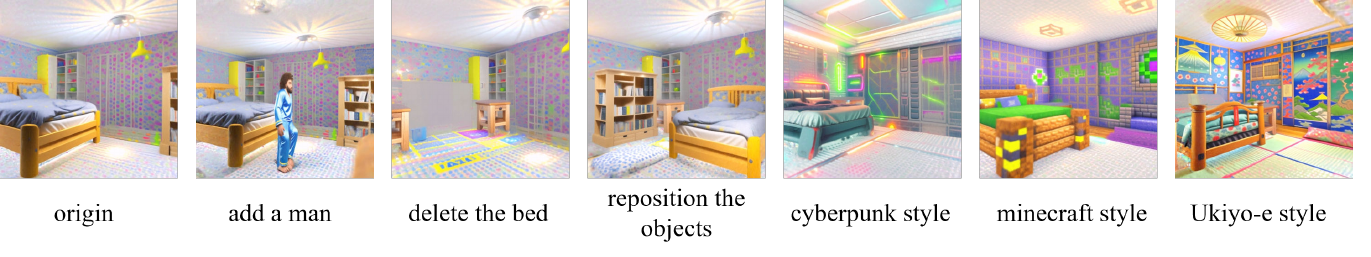}
\caption{ \sysname\ editing results.}
\label{fig:editiing}
\end{figure}
\vskip 0.1in \noindent\textbf{Implementation Details.}
We utilize GPT-4 as our LLM for scene prompt decomposition, Point-E~\cite{nichol2022point} generated sparse point clouds for the initial representation of objects, Stable Diffusion 2.1 as our 2D text-to-image model. The maximum number of iterations is set to 1,500 and 2,000 rounds for objects and the environment, respectively. The initial time interval value $m$, starts at 4 and decreases by 1 every 400 rounds. The number of rendering images in the reconstructive generation is set to 20. We tested \sysname\ and all the baselines on the same NVIDIA 3090 GPU for fair comparison.
\vskip 0.1in \noindent\textbf{Baselines.}
For the comparison of the text-to-3D scene generation, we use the current open-sourced SOTA methods Text2Room~\cite{hollein2023text2room}, Text2NeRF~\cite{zhang2024text2nerf}, ProlificDreamer~\cite{wang2024prolificdreamer}, and Set-the-Scene~\cite{cohen2023set} as the baselines.
For text-to-3D generation, we select open-source SOTA methods DreamFusion~\cite{poole2022dreamfusion}, Magic3D~\cite{lin2023magic3d}, DreamGaussian~\cite{tang2023dreamgaussian}, and LucidDreamer~\cite{liang2023luciddreamer} as the baselines (ProlificDreamer, DreamFusion and Magic3D were reimplemented by Three-studio~\cite{guo2023threestudio}).
\vskip 0.1in \noindent\textbf{Evaluation Metrics.}
We tested the generation time of each method~\cite{hollein2023text2room,zhang2024text2nerf,cohen2023set,wang2024prolificdreamer,poole2022dreamfusion,lin2023magic3d,tang2023dreamgaussian,liang2023luciddreamer}, compared the editing capabilities against their published papers, and did a 100-participant user study to score (out of 5) the quality, consistency, and rationality of the videos generated by each method  for 5 scenes (3 indoor, 2 outdoor) of 30 seconds.

\subsection{Qualitative Results}
Fig.~\ref{fig:scene_result} shows the comparison of \sysname\ in indoor and outdoor scenes with the SOTA methods\cite{hwang2023text2scene,zhang2024text2nerf,wang2024prolificdreamer,cohen2023set}. The upper images are rendered with the camera poses that appeared during the generation, and the lower ones with the randomly selected camera poses within the scene. We can see that Text2Room\cite{hollein2023text2room} and Text2NeRF\cite{zhang2024text2nerf} only produce satisfactory results under camera poses encountered during generation. Combined with the results from Fig.~\ref{fig:consistency}, it shows that \sysname\ achieves the best 3D consistency alongside commendable generation quality. For the comparison of generating single objects with text-to-3D methods, Fig.~\ref{fig:object_result} shows that our FPS can generate high-quality 3D representation following the text prompts in a short time.  DreamGaussian~\cite{tang2023dreamgaussian} is faster yet at the cost of too low generation quality. Please refer to the Supplementary Materials for additional images and video results.
\subsection{Quantitative Results}
Since baselines~\cite{wang2024prolificdreamer,zhang2024text2nerf,hollein2023text2room} are not able to generate objects in the environment independently, for a fair comparison we calculate the generation time of our environment generation stage. The left side of Tab.~\ref{tab:userstudy} shows that we have the shortest generation time for environments with editing capabilities, and the right side shows the user study, where \sysname\ is far ahead of ~\cite{wang2024prolificdreamer,cohen2023set,zhang2024text2nerf,hollein2023text2room} in terms of consistency and rationality with high generation quality.

\begin{figure}
\centering
\begin{subfigure}{0.23\linewidth}
\centering
\includegraphics[width=1\textwidth]{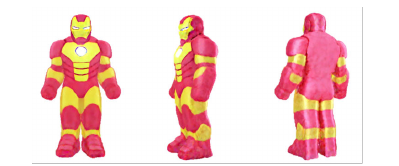}
\caption{SDS~\cite{poole2022dreamfusion}}
\end{subfigure}
\hfill
\begin{subfigure}{0.23\linewidth}
\centering
\includegraphics[width=1\textwidth]{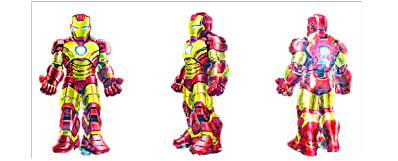}
\caption{DreamTime~\cite{huang2023dreamtime}}
\end{subfigure}
\hfill
\begin{subfigure}{0.23\linewidth}
\centering
\includegraphics[width=0.95\textwidth]{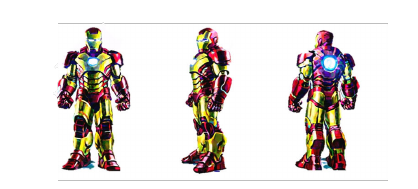}
\caption{MTS}
\end{subfigure}
\hfill
\begin{subfigure}{0.23\linewidth}
\centering
\includegraphics[width=0.95\textwidth]{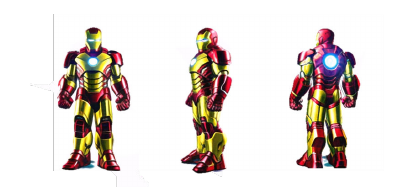}
\caption{FPS}
\end{subfigure}
\caption{The ablation results of different sampling strategies.}
\label{fig:gen_by_recon}
\end{figure}

\begin{figure}
\centering
\begin{subfigure}{0.3\linewidth}
\centering
\includegraphics[width=1\textwidth]{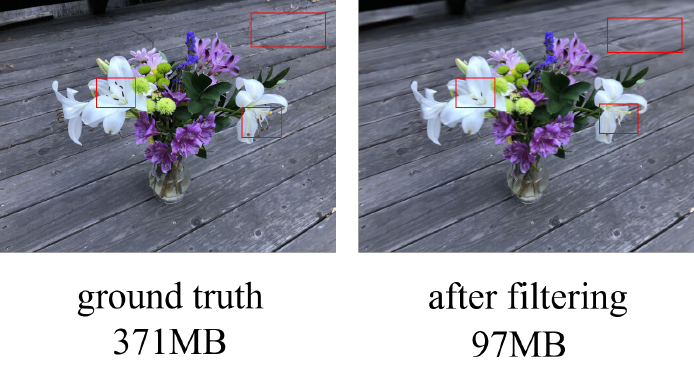}
\caption{data in NeRF-360\cite{mildenhall2021nerf}}
\end{subfigure}
\hfill
\begin{subfigure}{0.6\linewidth}
\centering
\includegraphics[width=\textwidth]{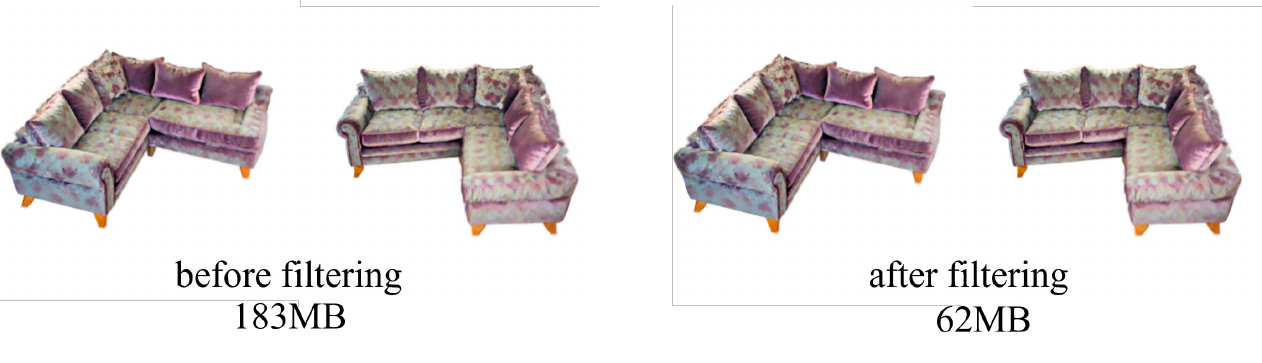}
\caption{data in generating process}
\end{subfigure}
\caption{Ablation results of 3D Gaussian filtering algorithm in reconstruction and generation tasks.}
\label{fig:faux_renderding}
\end{figure}

\begin{table}[t]
\centering

\setlength{\tabcolsep}{1.4mm}{
\begin{tabular}{c |c| c |c c  c}  \hline
          Method     & Time (hours) $\downarrow$&  Editing&\multicolumn{3}{|c}{User Study} \\
             &       &   &quality$\uparrow$&consistency$\uparrow$&rationality$\uparrow$    \\ \hline
        
        Text2Room~\cite{hollein2023text2room}  & 13.3 & \ding{55} & 2.93 & 2.57 & 2.60 \\ 
        Text2NeRF~\cite{zhang2024text2nerf} &  7.5&  \ding{55} & 3.05 & 2.71 & 2.98 \\ 
         ProlificDreamer~\cite{wang2024prolificdreamer}   & 12.0&\ding{55} & 3.48 & 3.19 & 2.95 \\ 
         Set-the-Scene~\cite{cohen2023set}   &1.5& \checkmark & 2.45 & 3.52 & 2.88 \\ \hline
         \textbf{Ours}   &\textbf{1.0}&   \checkmark & \textbf{3.92} & \textbf{4.24}& \textbf{4.05} \\  \hline
 \end{tabular} 
}

\caption{Quantitative Results of \sysname\ compared with baselines. $\uparrow$ means the more the better and $\downarrow$ means the lower the better.}   
\label{tab:userstudy}
\end{table}

\subsection{Scene Editing}
Fig.~\ref{fig:editiing} demonstrates the flexible editing capabilities of \sysname. \sysname\ can add or remove an object or ressign its position in the scene by adjusting the values of the object's affine components. When making these edits, we need to resample camera poses at the original and new locations of the object, re-optimizing towards the ground and surrounding directions. We can also change the text prompts to alter the style of the environments or the objects in the scene. This editing can be obtained after re-optimizing the coarse 3D representation. 
\subsection{Ablations}

%

We compared the effects of different sampling strategies via the generation results of a 3D object. Fig.~\ref{fig:gen_by_recon} shows the results optimized for 30 minutes under the prompt ``A DSLR photo of Iron Man''. As illustrated, multi-timestep sampling (MTS) forms better geometric structures and textures compared to the monotonically non-increasing sampling in \cite{huang2023dreamtime} and the SDS in \cite{poole2022dreamfusion}. Formation Pattern Sampling (FPS), building on top of MTS, employs a reconstruction method to create smoother and more plausible textures. The 3D Gaussian filtering we proposed is specifically designed for optimization tasks and can also be directly applied to reconstruction tasks~\cite{kerbl20233d}. Fig.~\ref{fig:faux_renderding}  compares the results of reconstruction and generation before and after compression using the Gaussian filtering algorithm. It can be seen that in the reconstruction task, our compression achieved 73.9\%, with the overall image slightly blurred and some details lost. In our generation task, however, the compression was 66.1\% without significant quality loss. Please find the ablation experiments on camera sampling and more comprehensive results in the Supplementary Materials.

%% file: sec/8_applications.tex
\section{Applications, Limitation and Conclusion}

\begin{figure}[t]
\centering
\begin{subfigure}{0.32\linewidth}
\centering
\includegraphics[width=1\textwidth]{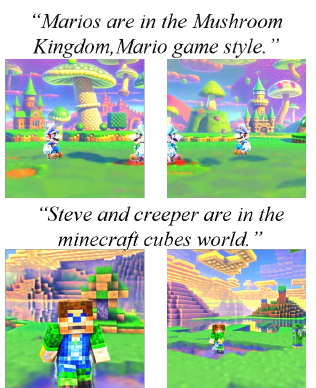}
\caption{gaming}
\end{subfigure}
\begin{subfigure}{0.32\linewidth}
\centering
\includegraphics[width=0.96\textwidth]{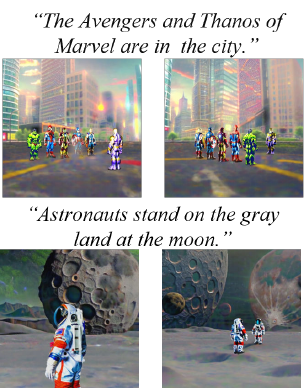}
\caption{film}
\end{subfigure}
\begin{subfigure}{0.32\linewidth}
\centering
\includegraphics[width=0.96\textwidth]{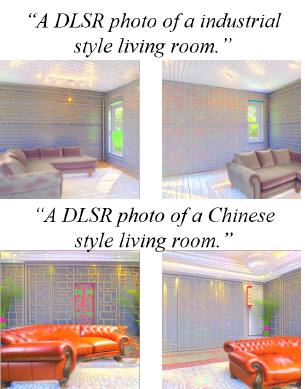}
\caption{architectural design}
\end{subfigure}
\caption{The potential applications of \sysname.}
\label{fig:application}
\end{figure}

\sysname\ can impact numerous industries as illustrated in Fig.~\ref{fig:application}. For instance, in VR gaming and the Metaverse, \sysname, integrated with LLMs, enables the capability to generate a fantasy scene merely from a user's description. Moreover, it allows for easy modification of object positions and the style of the scene through descriptive means. \sysname\ can also revolutionize film production by enabling rapid and customizable generation of intricate 3D scenes directly from scripts, significantly enhancing visual storytelling and reducing the reliance on physical sets and manual modeling. For architectural design, \sysname\ facilitates the generation of furniture in various styles, enabling users to flexibly design and arrange objects within a scene using layout diagrams.

\sysname\ currently cannot generate outdoor scenes as realistically close to actual environments as Text2Room~\cite{hollein2023text2room} and Text2NeRF~\cite{zhang2024text2nerf}. This is because their methods generate scenes through image inpainting, resulting in very limited observable camera poses and poor 3D consistency across the entire scene. In contrast, we sacrificed a degree of realism to ensure overall scene consistency. In future work, we plan to incorporate depth supervision to guide the generation of outdoor scenes with a style as realistic as indoor ones.

In summary, we introduce a novel text-to-3D scene generation strategy \sysname. By employing Formation Pattern Sampling (FPS), a camera sampling strategy, integrating objects and environments, we address the current issues of inefficiency, inconsistency, and limited editability in current text-to-3D scene generation methods. Extensive experiments have demonstrated that \sysname\ is a milestone achievement in 3D scene generation, holding potential for wide-ranging applications across numerous fields.
\par 
\par \textbf{Acknowledgements}: This work was supported by the National Key Research and Development Program of China (2022YFB3105405, 2021YFC3300502).

%% file: sec/9_supplement_arxiv.tex
\section{Discussions of Current Methods}

\noindent\textbf{Inpainting-based methods} employ text-to-image inpainting for scene generation~\cite{zhang2024text2nerf,hollein2023text2room,ouyang2023text2immersion}. They first initialize an image and then partially mask it to represent a view from an alternate angle. Leveraging pretrained image inpainting models like Stable Diffusion~\cite{rombach2022high}, alongside depth estimation, they reconstruct the occluded image segments and deduce their depths, iteratively composing the entire scene through depth and image alignment. While such methods can achieve good visual effects at the camera positions (e.g., the center of the scene) during the generation process, they encounter significant limitations in visible range. Venturing beyond the predefined camera areas used during generation precipitates scene degradation, as exemplified in Fig.~\textcolor{red}{4a} of the paper, underscoring a deficiency in scene-wide 3D consistency. Concurrently, this method's tendency to replicate certain environmental objects, like the proliferation of sofas in a living room scenario, highlights issues with logical scene composition. 




\noindent\textbf{Combination-based methods} also employ a combination approach to construct scenes~\cite{cohen2023set,zhang2023scenewiz3d}. However, they face challenges including low generation quality and slow training speeds. Moreover, \cite{zhang2023scenewiz3d} utilizes various 3D representations (such as NeRF$+$DMTet) to integrate objects and scenes, increasing the complexity of scene representation and thus limiting the number of objects that can be placed within the scene (2-3 objects), impacting their applicability. In contrast, \sysname's Formation Pattern Sampling (FPS) can generate high-quality 3D content in a very short time, using a single 3D representation to compose the entire scene, allowing for more than 20 objects to be placed within the scene. This underscores \sysname's remarkable superiority.



\noindent\textbf{Objects combination  methods} do not take the environmental context into account, focusing solely on whether the combination of objects is logical~\cite{zhou2024gala3d,vilesov2023cg3d,lin2023componerf}. They generate a simple assembly of objects rather than a complete scene. We believe that the approach to scene composition should be more diverse and offer greater controllability.


\par \sysname\ demonstrates a significant advantage by efficiently, consistently, and flexibly generating 3D scenes, showcasing a substantial superiority over the aforementioned methods. 
%
\section{Additional Implementation Details}
The overall generation process of \sysname\ is shown in Algorithm~\ref{alg:DreamScene}.

\subsection{Rendering and Training Settings}

We render images with a size of 512×512 for optimization. During the optimization process, we do not reset the opacity, to maintain the consistency of the optimization during the training process and avoid gradient disappearance due to opacity reset.






\begin{algorithm}
\caption{\sysname}
\label{alg:DreamScene}
\begin{algorithmic}[1] 
\State $Y \rightarrow y_1,y_2,...,y_N,y_e$;
\For{$n = [1,2,...,N,e]$}
    \If{$n$ is not $e$}
        \State Initialize 3D Gaussian of $obj_n$ 
    \Else
        \State Initialize 3D Gaussian of environment
    \EndIf
    \For{$iter = [0,1,...,max\_iter]$}
        \If{$n$ is not $e$}
            \State Sample camera pose $c$
        \Else
            \State Sample camera pose $c$ follow strategy in Sec.~\ref{camera_sampling}
        \EndIf
        \State $x_0 = g(\theta,c)$
        \State $T_{end} = (1-\frac{iter}{max\_iter}) timesteps$
        \For{$i = [1,2,...,m]$}
            \State $t_i = T_{end}\cdot random(\frac{i-1}{m},\frac{i}{m})$
            \State $x_i=DDIM_(x_{i-1},i)$
            \State $\epsilon_{\phi}(x_{t_i}; y_n, t_i) = $U-Net$(x_{t_i},y_n,t_i)$
            \State $\epsilon_{\phi}(x_{t_i}; \emptyset, t_i) = $U-Net$(x_{t_i},\emptyset,t_i)$ 
        \EndFor
        \State $\nabla_{\theta}\mathcal{L}_{\text{MTS}}(\theta) = \mathbb{E}_{t,\epsilon,c}\left[\sum\limits_{i=1}^{m} w(t_i)(\epsilon_{\phi}(x_{t_i}; y_n, t_i) - \epsilon_{\phi}(x_{t_i}; \emptyset, t_i))\frac{\partial g(\theta,c)}{\partial \theta}\right]$
        \State Update $\theta$
        \If{$iter\%compress\_iter=0$}
        \State $Score_k = \sum_{j=1}^{H\times W \times M} \frac{V(k)}{D(r_j,k)^2\times maxV(r_j)} $
        \State $Sort(Score_k)$
        \State Delete last $z$ 3D Gaussians
        \EndIf
    \EndFor
    \If{$n$ is $e$}
        \State Save 3D Gaussian Representation of the Scene
        \State break
    \EndIf
    \State Save 3D Gaussian Representation $obj_n$ of text $y_n$
    \State $world(x) = r\cdot s \cdot obj_n(x) + t$
    \State Add $obj_n$ to the Scene
\EndFor
\end{algorithmic}
\end{algorithm}

\subsection{Camera Sampling Strategy}
\label{camera_sampling}
This section outlines a three-stage camera sampling strategy for crafting both outdoor and indoor scenes. The process is as follows:
\subsubsection{Outdoor}
\begin{itemize}
\item In the first stage, we freeze the parameters of the ground and objects, focusing solely on optimizing the surrounding environments without rendering the objects. During this phase, we sample cameras near the center of the scene, the camera's pitch angle is set between 80 to 110 degree. After reaching 70\% of the iterations of this stage, we select four camera poses for multi-camera sampling at each later iteration. These four cameras, all directed towards the same direction, are positioned on either side of the scene center at distances of either 1/4 or 1/2 of the radius, ensuring the environments achieve satisfactory visual effects across various distances.


\item In the second stage, we freeze the parameters of the surrounding environments and objects, and focus solely on optimizing the ground without rendering objects. We sample four camera poses at each iteration, akin to the sampling strategy in the later part of the first stage. We adjust the pitch angle range to 85$\sim$95 degree, which can reduce the occurrence of a singular ground or environment in the rendered images, thereby enhancing the overall scene generation outcome.


\item In the third stage, we optimize both the surrounding environments and the ground, and render objects into the scene to achieve a harmonious and unified effect. We integrate the camera positions used in the previous two stages, ensuring that areas within the scene are evenly and comprehensively covered, thereby attaining a consistent reconstruction result.
\end{itemize}

\subsubsection{Indoor}
\begin{itemize}

\item In the first stage, we freeze the ground and object parameters and \textbf{render} the objects into the scene. We primarily sample camera poses around the center of the scene and set the radius as large as possible to encompass all objects and thereby minimize the multi-head problem. At this stage, the pitch angle range of cameras is set to 75$\sim$115 degree. After the same iterations at outdoor settings, we sample camera poses around the objects to reduce the impact of object occlusion on the environment.


\item In the second stage, we freeze the environment parameters and begin optimizing the ground parameters. The indoor camera sampling strategy remains largely unchanged, but we adjust the pitch angle range to 45$\sim$90 degree to ensure coverage of the ground. Additionally, we increase the camera sampling around objects and from the center to the periphery of the scene, thereby enhancing the integration between the ground and the walls.




\item In the third stage, we optimize both the surrounding environments and the ground, and render objects into the scene in the same way as outdoor's.

\end{itemize}

Our indoor and outdoor camera sampling strategies correspond to typical bounded and unbounded scenes, respectively. For bounded scenes, we focus on ensuring consistency across various orientations. Hence we pay more attention to integrating objects with environments to avoid illogical layouts caused by generating excessive objects in the environment. For unbounded scenes, our primary concern lies with maintaining scene-wide consistency across varied distances. Therefore, we employ two different strategies for scene generation.

\section{Additional Results}

\subsection{Qualitative Results}
\begin{figure}
\centering
\includegraphics[width=1\textwidth]{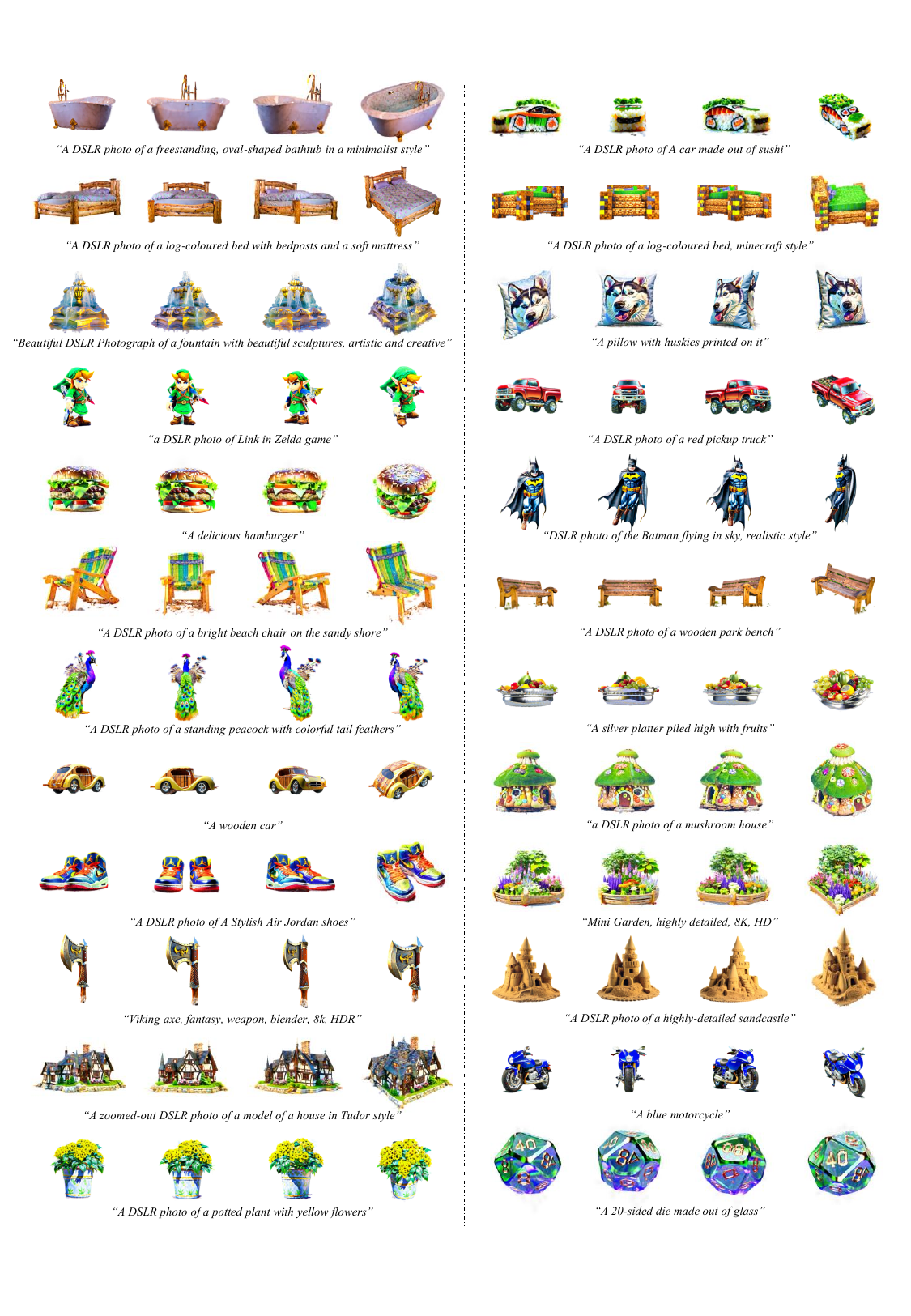}

\caption{More object generation results of \sysname.
}
\label{fig:supp_object_result}
\end{figure}

\begin{figure}
\centering
\includegraphics[width=1\textwidth]{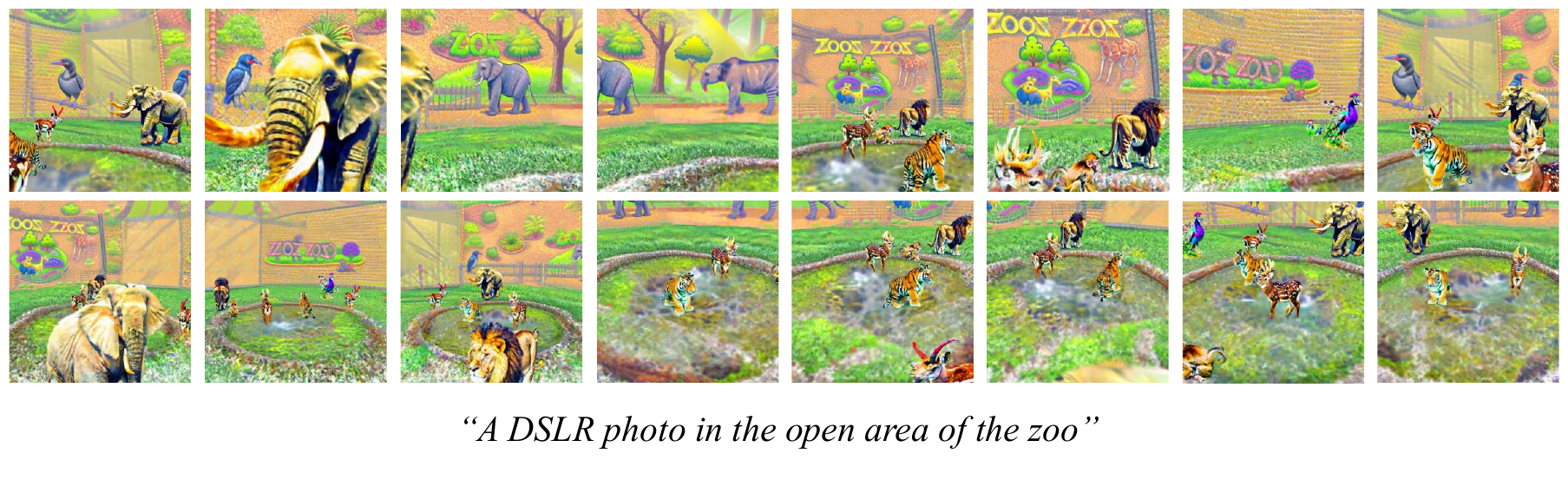}
\includegraphics[width=1\textwidth]{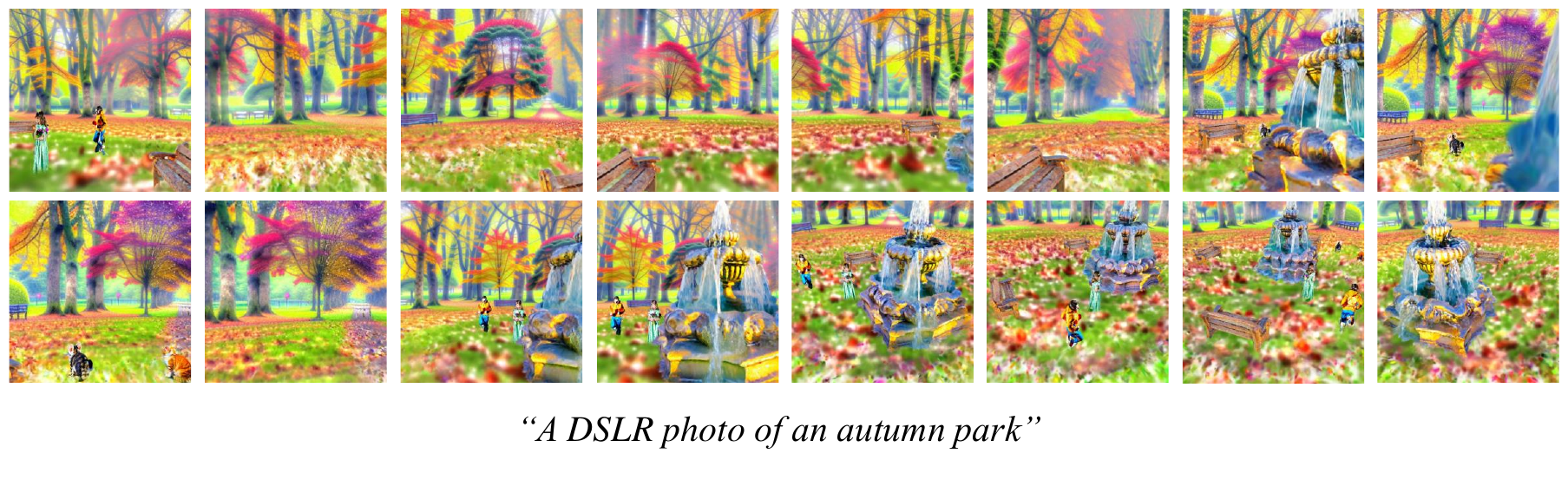}
\includegraphics[width=1\textwidth]{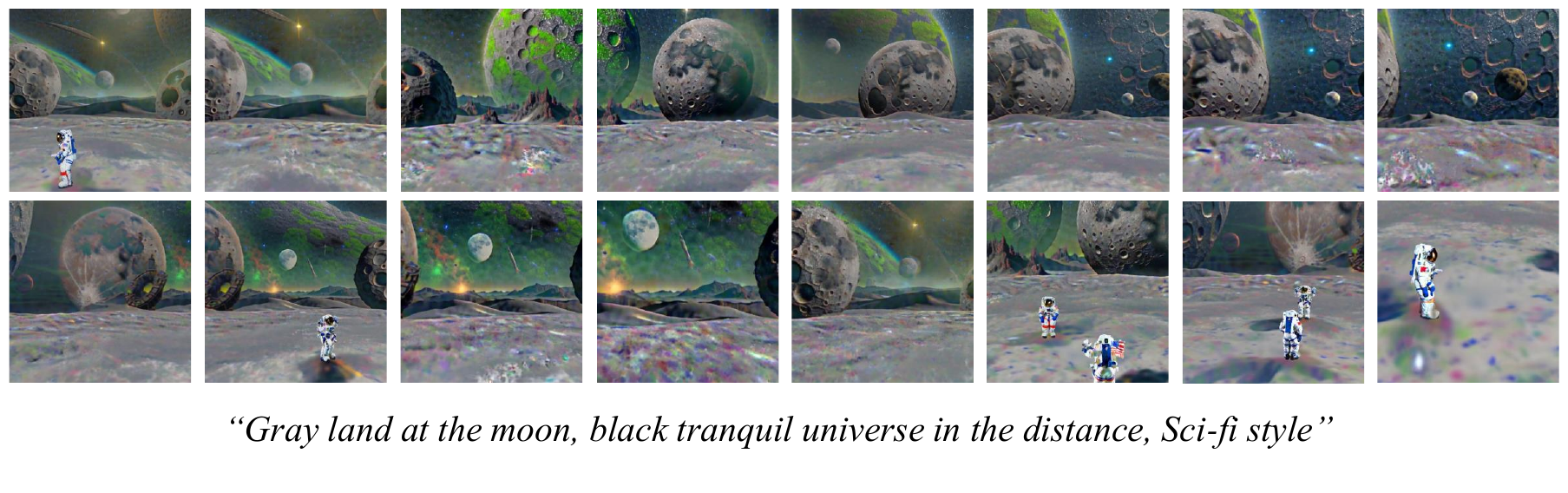}
\includegraphics[width=1\textwidth]{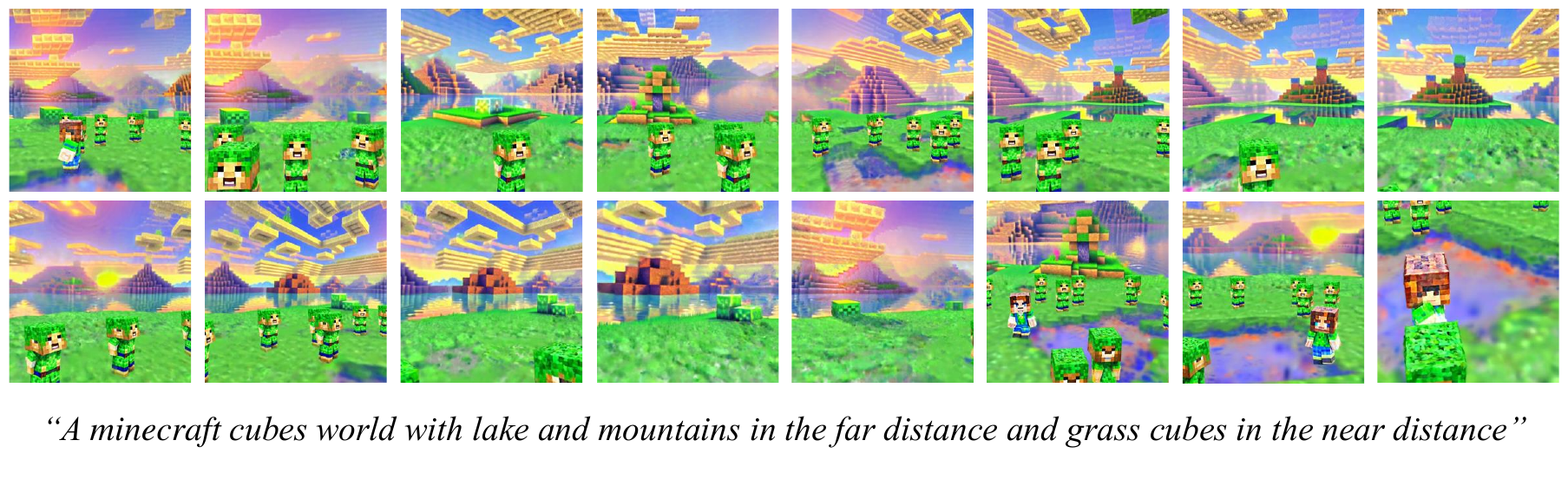}
\caption{More outdoor scene generation results of \sysname.}
\label{fig:supp_outdoor_result}
\end{figure}

\begin{figure}
\centering
\includegraphics[width=1\textwidth]{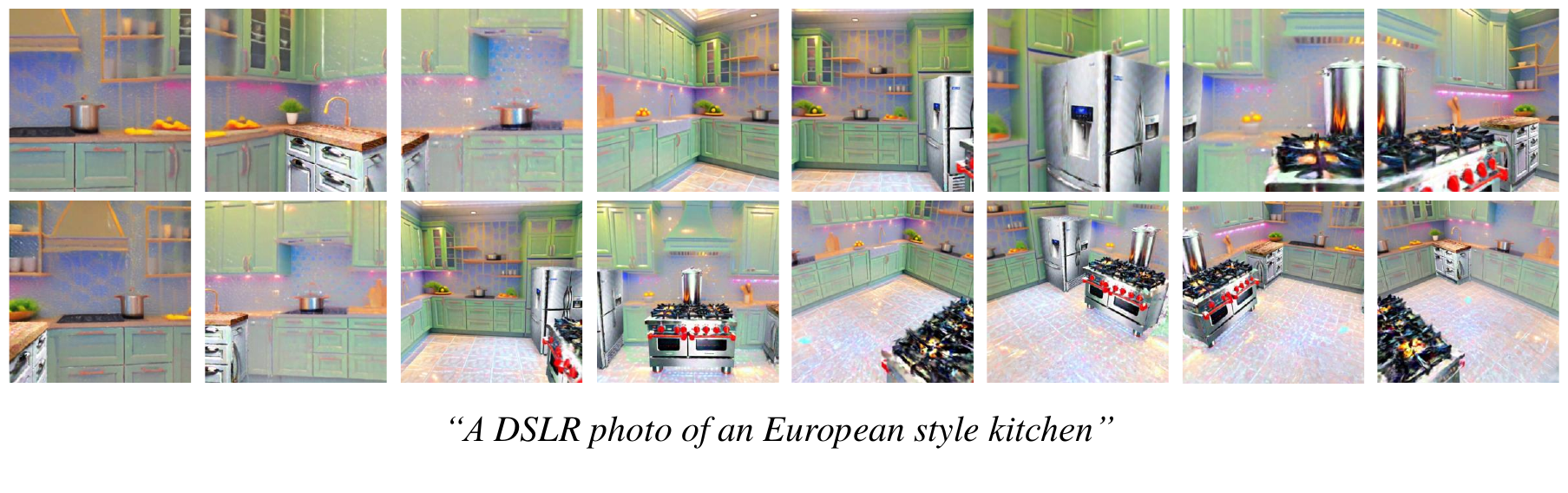}
\includegraphics[width=1\textwidth]{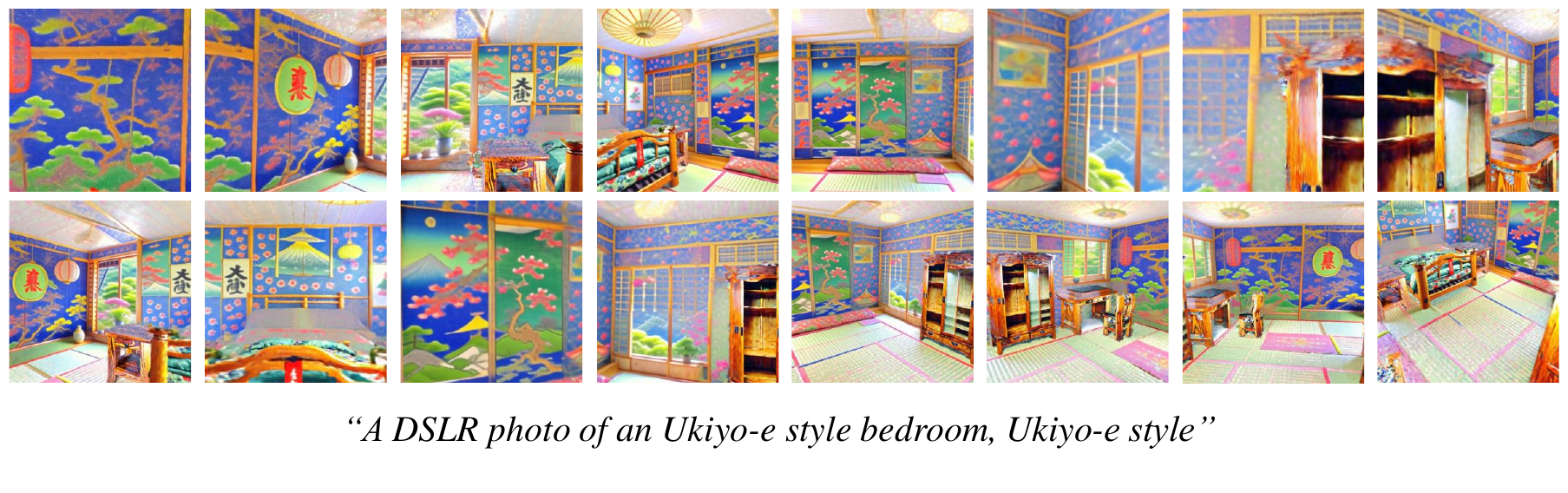}
\includegraphics[width=1\textwidth]{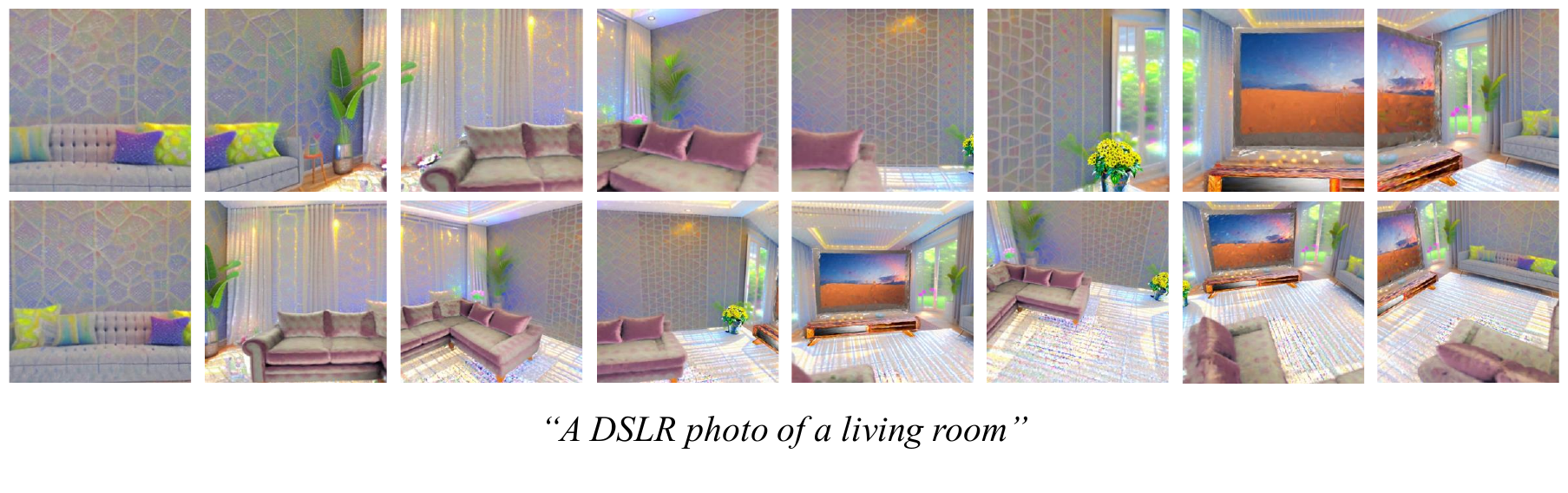}
\caption{More indoor scene generation results of \sysname.}
\label{fig:supp_indoor_result}
\end{figure}

More qualitative results of FPS are shown in Fig.~\ref{fig:supp_object_result}.
More qualitative results of scene generation are shown in Fig.~\ref{fig:supp_outdoor_result} and Fig.~\ref{fig:supp_indoor_result}.



\begin{figure}
\centering
\begin{subfigure}{0.3\linewidth}
\centering
\includegraphics[width=1\textwidth]{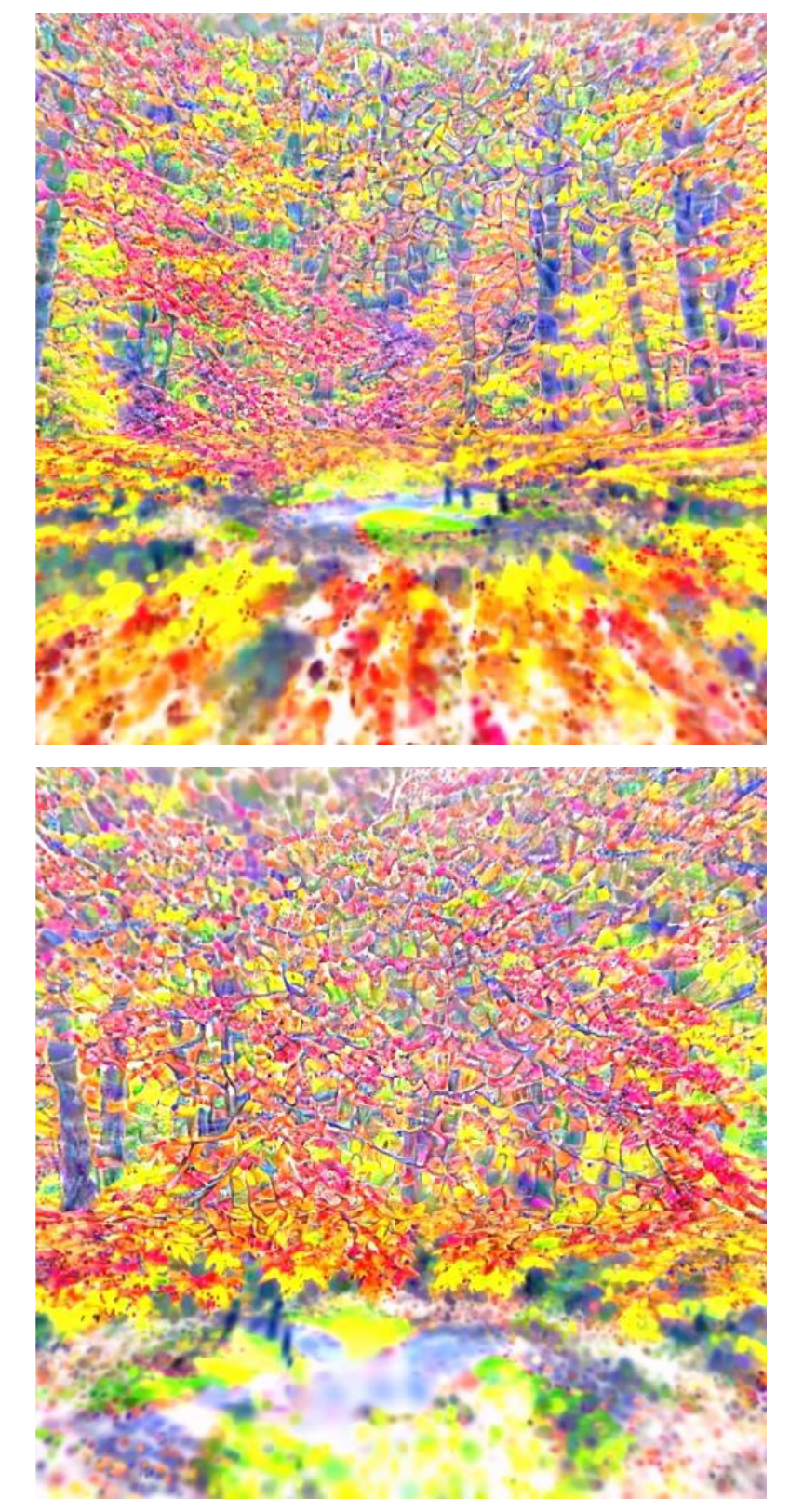}
\caption{Randomly camera sampling}
\end{subfigure}
\hfill
\begin{subfigure}{0.3\linewidth}
\centering
\includegraphics[width=1\textwidth]{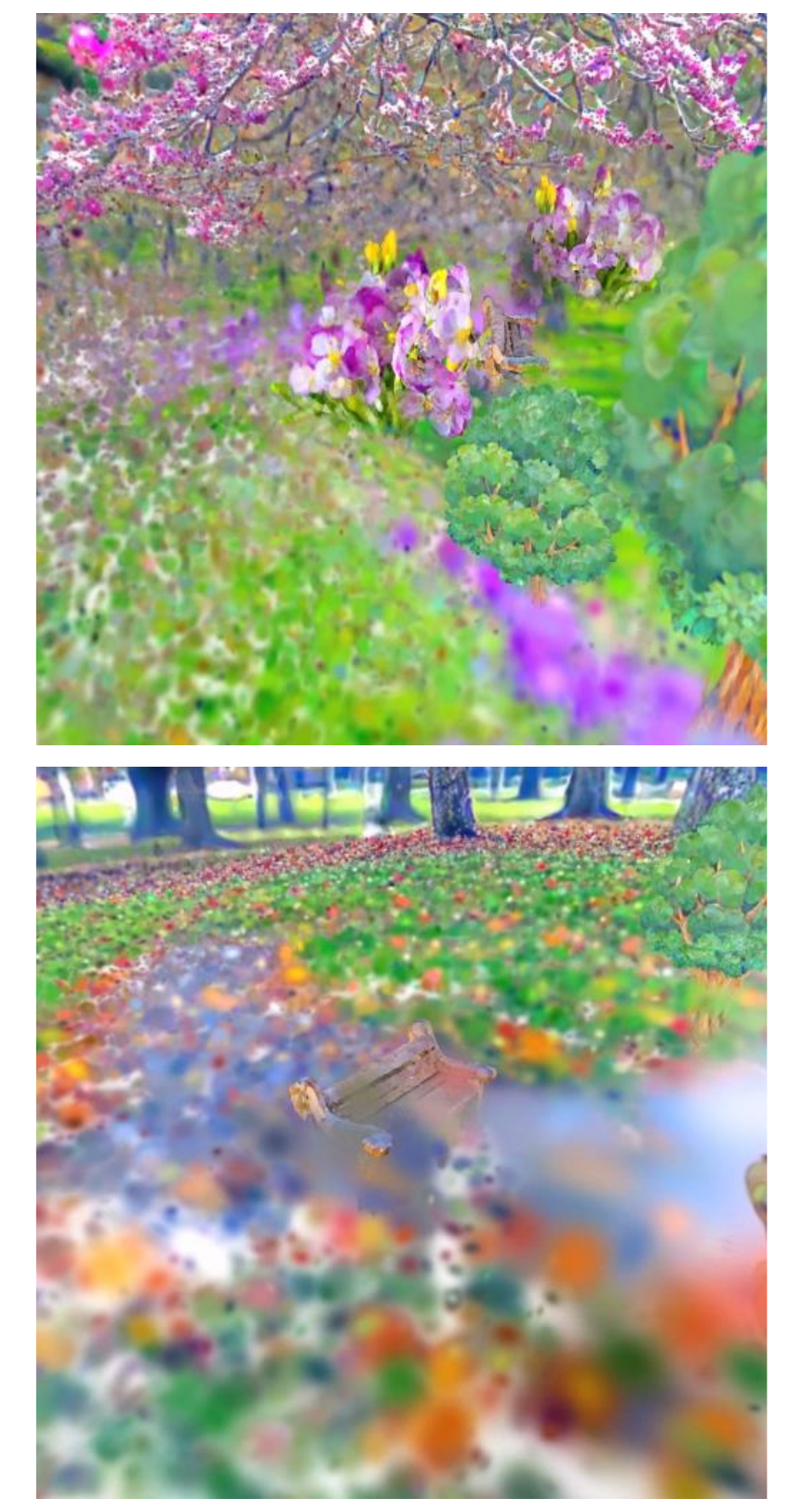}
\caption{No distinction between environment and ground}
\end{subfigure}
\hfill
\begin{subfigure}{0.3\linewidth}
\centering
\includegraphics[width=1\textwidth]{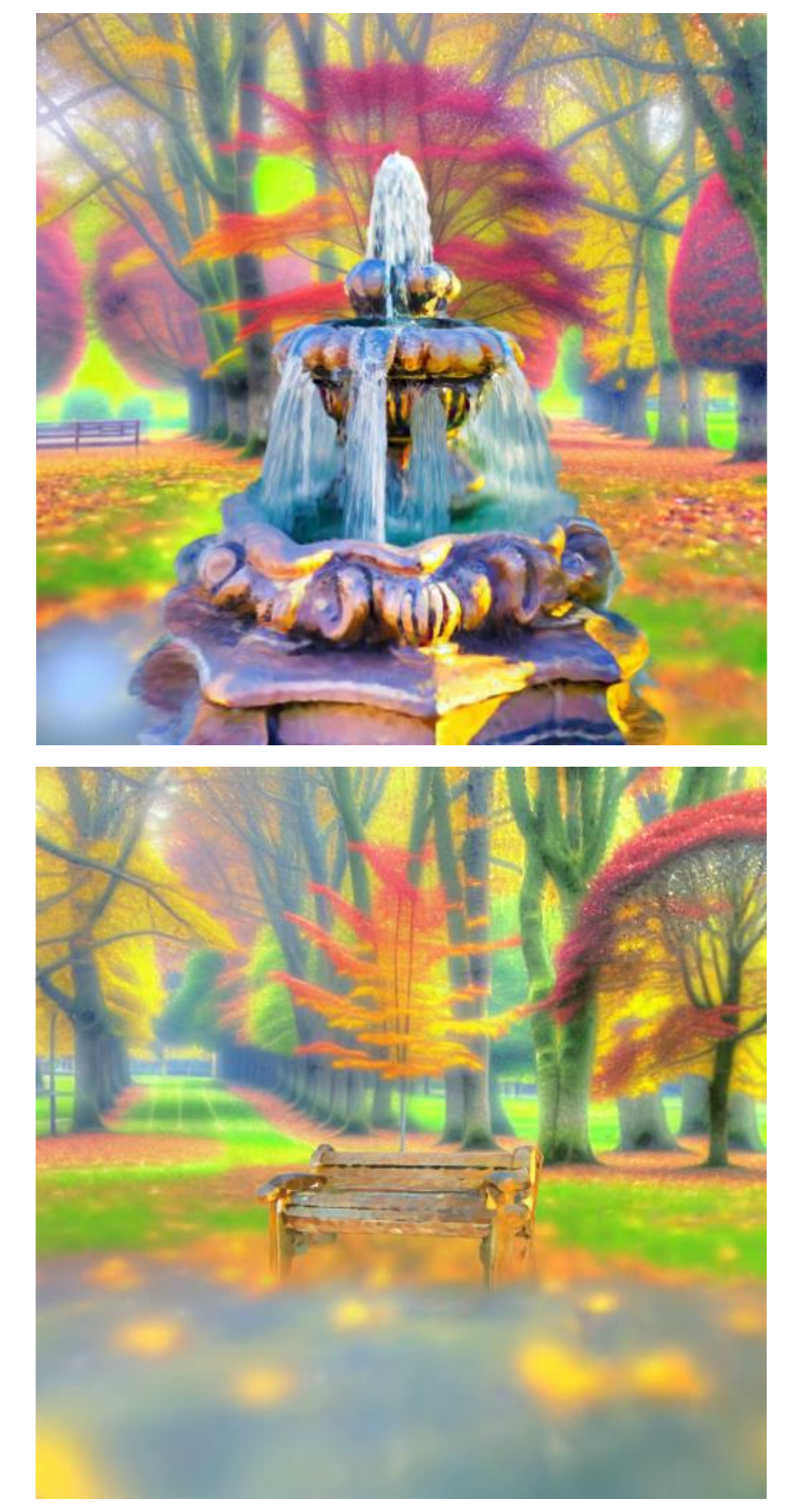}
\caption{\sysname\ three-stage camera sampling strategy}
\end{subfigure}
\vspace{-5pt}
\caption{The ablation results of different camera sampling strategies.}
\label{fig:sampling_ablation}
\end{figure}

\subsection{Ablation Study}
Fig.~\ref{fig:sampling_ablation}(a) depicts a scene generated by randomly sampling cameras within the scene. Due to the difficulty in ensuring consistency of multi-angle views at the same location, the optimization process often tends to collapse. Fig.~\ref{fig:sampling_ablation}(b) utilizes a strategy that initiates from the center to the surroundings, where the environment and ground are not differentiated. It can be observed that while scene consistency improves, the connection between the ground and the scene is poorly generated, and the ground is prone to coarse Gaussian points. Fig.~\ref{fig:sampling_ablation}(c) employs our three-phase strategy, enhancing the generation quality while ensuring the consistency of the surrounding environment and ground.